\documentclass[10pt,twocolumn,letterpaper]{article}

\usepackage[pagenumbers]{cvpr} 

\usepackage{graphicx}
\usepackage{amsmath}
\usepackage{amssymb}
\usepackage{booktabs}
\usepackage{enumitem}
\usepackage{fancyhdr}

\usepackage{amsthm}
\usepackage{pifont}
\usepackage[dvipsnames]{xcolor}
\usepackage{colortbl}
\usepackage{pgfplots}
\pgfplotsset{compat=1.18}
\usepackage{subcaption}
\usepackage{multirow}

\usepackage[pagebackref,breaklinks,colorlinks]{hyperref}

\usepackage[capitalize]{cleveref}
\crefname{section}{Sec.}{Secs.}
\Crefname{section}{Section}{Sections}
\Crefname{table}{Table}{Tables}
\crefname{table}{Tab.}{Tabs.}

\newcommand{\cmark}{\ding{51}}%
\newcommand{\xmark}{\ding{55}}%

\definecolor{LightCyan}{rgb}{0.7,1,1}
\definecolor{LightYellow}{rgb}{1,1,0.7}
\definecolor{LightOrange}{rgb}{1,0.8,0.5}

\newcolumntype{A}{>{\columncolor{LightCyan}}r}
\newcolumntype{B}{>{\columncolor{LightYellow}}r}
\newcolumntype{D}{>{\columncolor{LightOrange}}r}

\def\paperID{11759} 
\def\confName{CVPR}
\def\confYear{2025}

\begin{document}

\title{ProbPose: A Probabilistic Approach to 2D Human Pose Estimation}


\author{Miroslav Purkrabek \hspace{0.05cm} and \hspace{0.05cm} Jiri Matas
\vspace{0.1cm}\\
Visual Recognition Group\\
Department of Cybernetics\\
Faculty of Electrical Engineering\\
Czech Technical University in Prague\\
{\tt\small purkrmir@fel.cvut.cz}
}



\maketitle

\begin{abstract}

Current Human Pose Estimation methods have achieved significant improvements. However, state-of-the-art models ignore out-of-image keypoints and use uncalibrated heatmaps as keypoint location representations.
To address these limitations, we propose ProbPose,
which predicts for each keypoint: a calibrated
probability of keypoint presence at each location in the activation window, the probability of being outside of it,
and its predicted visibility.
To address the lack of evaluation protocols for out-of-image keypoints, we introduce the CropCOCO dataset and the Extended OKS (Ex-OKS) metric, which extends OKS to out-of-image points.
Tested on COCO, CropCOCO, and OCHuman, ProbPose shows significant gains in out-of-image keypoint localization while also improving in-image localization through data augmentation. Additionally, the model improves robustness along the edges of the bounding box and offers better flexibility in keypoint evaluation.
The code and models are available on the \href{https://mirapurkrabek.github.io/ProbPose/}{project website}\footnote{\url{MiraPurkrabek.github.io/ProbPose/}} for research purposes.

\end{abstract}

\section{Introduction}
\label{sec:intro}

Human pose estimation (HPE) has seen significant progress, achieving robust performance on standard datasets.
The most successful models like ViTPose \cite{ViTPose} typically use a top-down, heatmap-based approach, where heatmaps serve as dense representations of keypoint locations across an image.

In this paper, we address two critical limitations of top-down HPE methods: (1) all previous methods overlook out-of-image keypoints during training and, more importantly, ignore them in evaluation, and (2) dense heatmaps restrict their use to simple point estimates.
While seemingly separate, these issues are interrelated and can be addressed through an appropriate choice of representation and loss function.

Top-down methods localize keypoints within a specific region called the activation window.
This activation window is the part of the image mapped to the predicted heatmap.
The activation window often matches the model input and it may exceed image boundaries due to preprocessing steps, such as maintaining aspect ratios or enlarging bounding boxes.

\begin{figure}[t]
    \centering
    \includegraphics[width=0.495\linewidth]{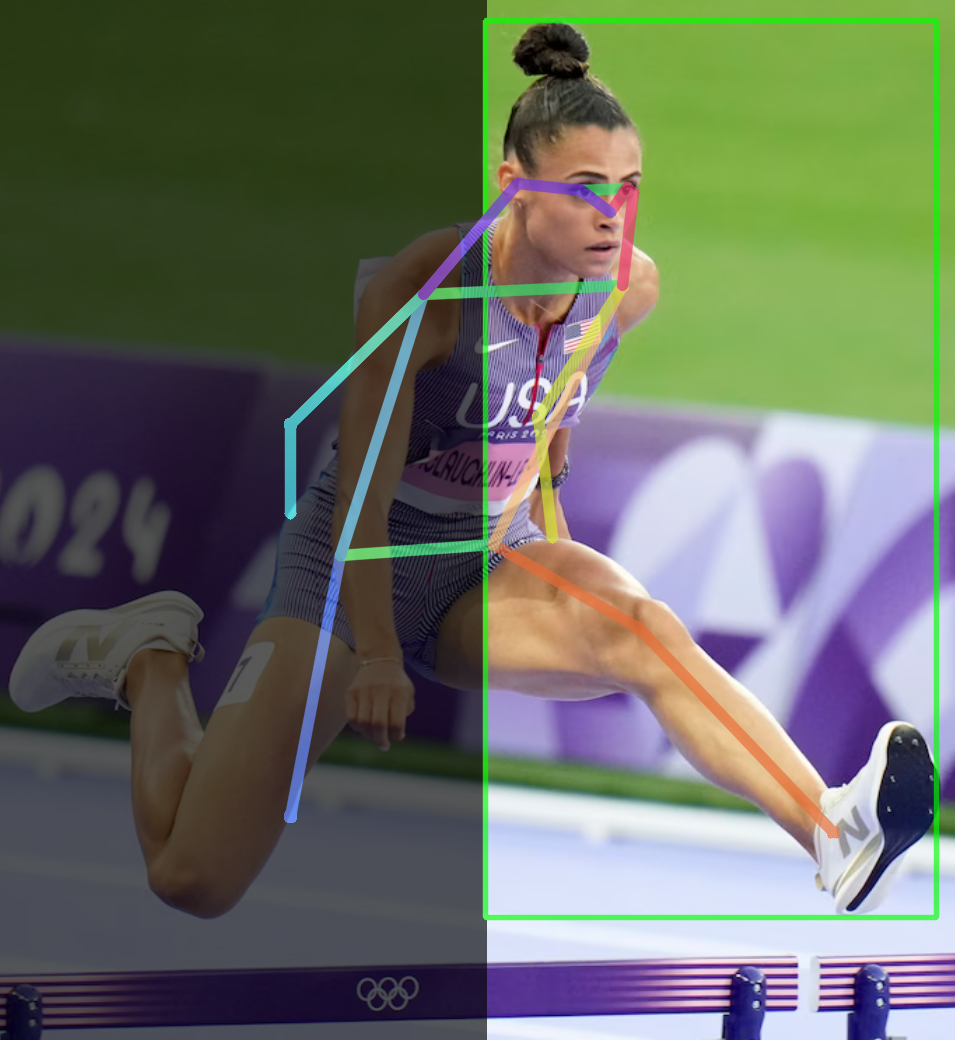}
    \hfill
    \includegraphics[width=0.495\linewidth]{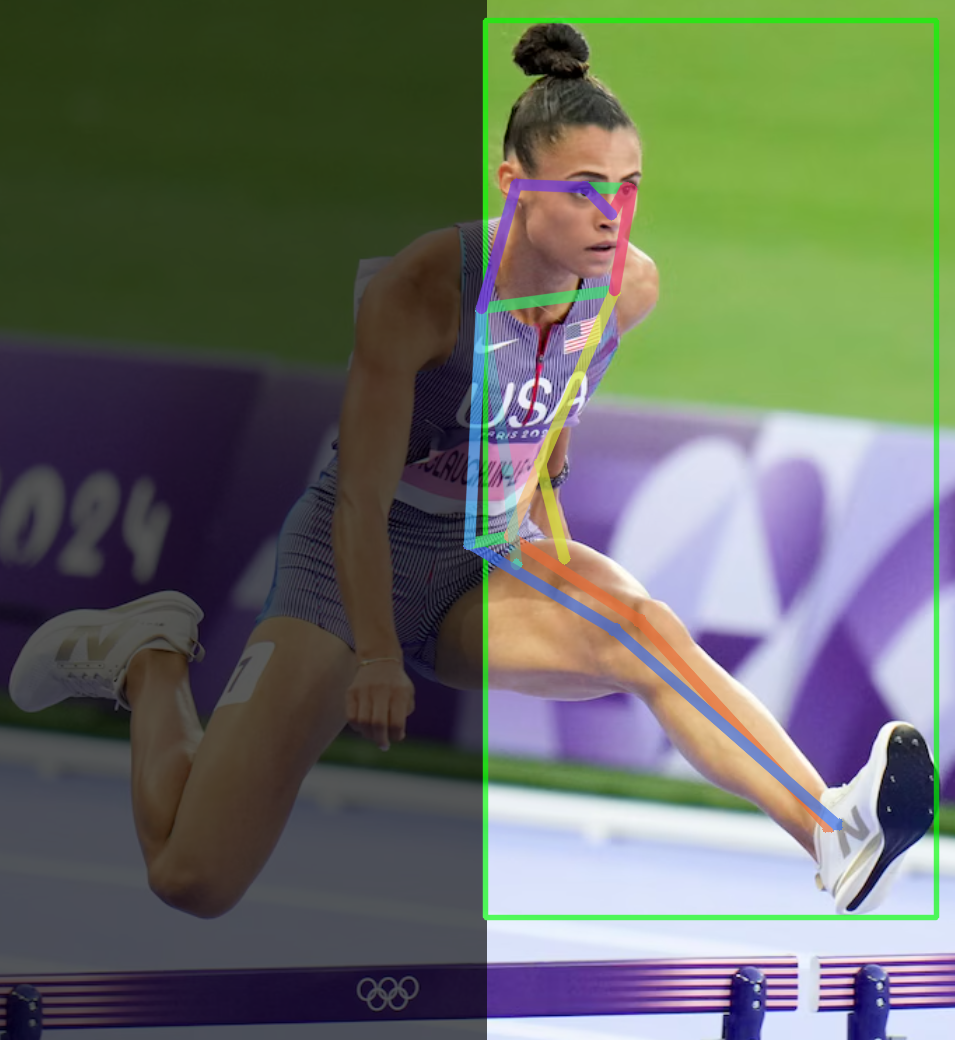}
    \caption{
   ProbPose (left) vs. ViTPose \cite{ViTPose} (right) output on  a cropped image with the dark part removed. 
   The bounding box, on which the pose estimators operate, is detected by YOLOX-x \cite{YOLOX}. 
    ProbPose estimates keypoints outside of the cropped image much better, including the right leg which is wrongly aligned with the left leg by ViTPose.}
    \label{fig:ooi-comparison}
\end{figure}

Under challenging conditions, such as heavy occlusion or cropped images, keypoints may fall outside the activation window.
Current methods do not address this issue and simply ignore out-of-window keypoints during training.
Even more concerning, evaluation metrics such as OKS (Object Keypoint Similarity) and PCK (Percentage of Correct Keypoints) assess only in-image keypoints.
This lack of accountability often leads models to predict keypoints inaccurately on other joints when the actual keypoint is not visible in the image.
For example, as shown in \cref{fig:ooi-comparison}, the left leg is mistakenly assigned to the same joints as the right leg.
Furthermore, current top-down models do not provide any indication of whether a keypoint is inside or outside the activation window, and models are not penalized for attempting to localize every keypoint.

A second challenge in existing HPE methods lies in the reliance on heatmaps as the keypoint representation.
Heatmaps provide dense representation of keypoint locations.
However, the standard approach of generating target heatmaps with a fixed Gaussian sigma and training with mean square error (MSE) forces the model to output heatmaps with fixed sigmas as well, limiting the accuracy of these estimates.

Decoding heatmaps to obtain a point estimate typically involves taking the argmax, often refined with UDP decoding \cite{UDP}.
While this approach is effective for identifying the most likely pixel, it is optimal only when the goal is to classify a single correct pixel for each keypoint.
However, practical applications demand more flexible interpretations of keypoint locations.
For example, an application might require an estimate of the smallest area where a keypoint is located with at least 95\% probability, or, as in human-robot interaction, identify areas where the model has high uncertainty to ensure safe interactions.
Heatmaps, which are prone to overconfidence, are not designed to indicate uncertainty or convey “don’t know” outcomes, even when Bayes risk is high, an essential factor in safety-critical applications.

To address these limitations, we propose ProbPose, a model with multiple outputs that fully describe each keypoint.
ProbPose predicts (1) calibrated \textit{presence probabilities} indicating whether the keypoint lies within the activation window, (2) keypoint's location within the activation window, (3) quality estimate of the localization and (4) visibility.
Moreover, instead of using traditional heatmaps, we employ probability maps, a more versatile tool for keypoint localization.

Probability maps differ from heatmaps in several ways: they contain values between 0 and 1 and are normalized to sum to 1, providing calibrated probabilities.
Unlike heatmaps, which assume a fixed Gaussian shape, probability maps adapt their shape based on the data, without forcing any assumptions.
Additionally, instead of decoding using a simple argmax, we compute the expected OKS for each pixel and select the location that maximizes this score, an approach we term expected OKS maximization.
Probability maps are trained using an OKS-based loss modified for dense predictions, formulated as an expected risk minimization problem.


Since current datasets and metrics lack evaluation protocols for out-of-image keypoints, we introduce a new dataset, CropCOCO, comprising cropped images from COCO, and the Extended OKS (Ex-OKS) metric, which builds on OKS to evaluate out-of-image keypoints and presence probability.

In summary, our main contributions are:
\begin{enumerate}
    \item A concept of \textbf{presence probability} that keypoints is in the activation window, distinct from confidence, which measures the model’s trust in its own estimate.

    \item \textbf{Model for out-of-image keypoints} that can localize keypoints within the activation window, even beyond the image boundaries, or predict presence probability for out-of-window keypoints.

    \item \textbf{OKSLoss adapted for dense predictions} in risk minimization formulation

    \item We show that \textbf{cropping-based data augmentation} improves out-of-image keypoint localization, supports presence probability estimation, and enhances in-image localization, similar to Hide\&Seek \cite{InformationDropping}

    \item New \textbf{CropCOCO dataset} and \textbf{Ex-OKS} evaluation metric for more realistic assesment of real-world performance
    
\end{enumerate}

\section{Related work}
\label{sec:related}

There are three main approaches to 2D human pose estimation: top-down, bottom-up, and single-stage.
Single-stage \cite{PETR, POET, CID, AdaptivePose++} methods directly predict individuals and their skeletons in an image.
Bottom-up methods \cite{OpenPose, DEKR} detect all keypoints first and then group them into individual skeletons.
The top-down methods \cite{HRNet, ViTPose, SWIN, PVTv2, HRFormer}, despite their challenges, remain the most successful.
They use human detectors to process each bounding box independently, providing the best performance on most datasets.

Among top-down methods, heatmap-based approaches are the most popular.
Unlike regression-based methods \cite{DeepPose, PoseRegression}, which directly predict 2D coordinates, heatmap-based methods \cite{ViTPose, SWIN, HRNet} predict a heatmap with the same aspect ratio as the input.
The keypoints are then estimated as the maximum of the heatmap.
Heatmaps are more robust and easier to train than regression-based methods, although efforts have been made to narrow the gap between the two approaches \cite{IntegralPose, IntegralPose2}.

\begin{figure}[t]
  \centering
  \begin{subfigure}{\linewidth}
    \centering
    \includegraphics[width=0.495\linewidth]{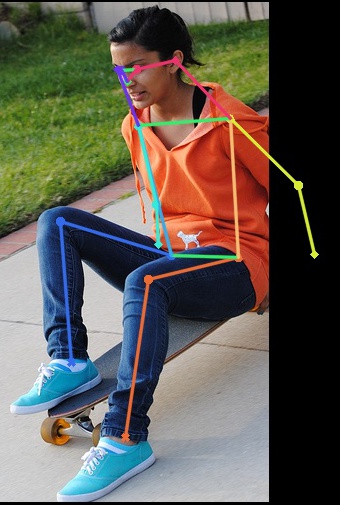}
    \hfill
    \includegraphics[width=0.495\linewidth]{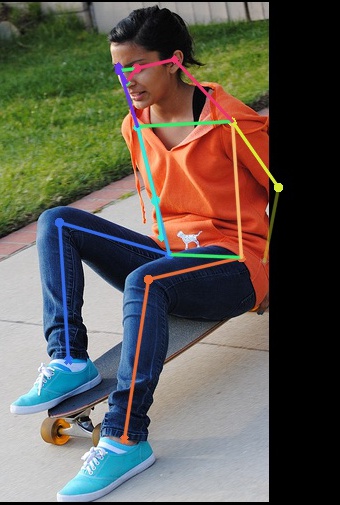}
  \end{subfigure}

  
  \begin{subfigure}{\linewidth}
    \centering
    \includegraphics[width=0.495\linewidth]{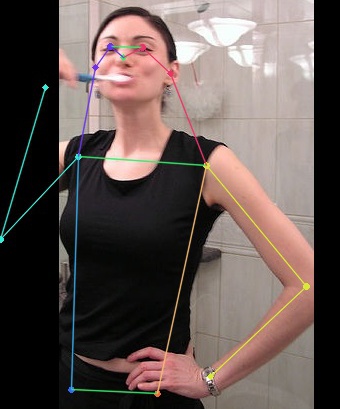}
    \hfill
    \includegraphics[width=0.495\linewidth]{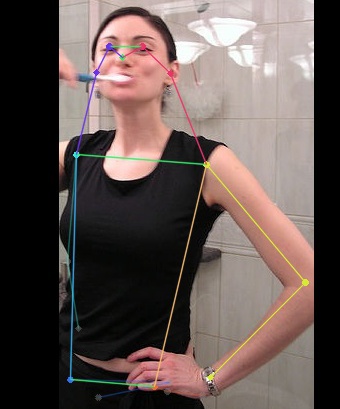}
  \end{subfigure}


    \caption{
    COCO (first row) and the proposed CropCOCO (second row) dataset comparison. 
    Compared with ViTPose-s (right), ProbPose (left) improves localization outside the image and handles occlusion better (second row).
    It also pushes keypoints to correct positions, even near the image border (third row).
    ProbPose estimates points well outside the original image.
    }
    \label{fig:OoI-examples}
  
\end{figure}

A key development in heatmap-based methods is UDP (unbiased data processing) \cite{UDP}, which improves coordinate encoding and decoding.
Heatmaps are typically defined as 2D Gaussians with a fixed sigma centered on the keypoint.
Subpixel decoding allows for smaller heatmaps, reducing computational costs.
The maximum value of the heatmap, called the "confidence," is often used for three tasks: visibility estimation, quality estimation, and in/out-of-image decisions.
Visibility is sometimes outsourced to specialized heads, quality prediction could be improved by OKS prediction, as shown in \cite{Calibration}.
In/out decisions (later defined as \textit{presence probability} in this paper) are not explicit, as they are not needed for OKS evaluation, but are important for visualization and real-world applications.

Most heatmap-based methods are trained with MSE loss, which encourages the output heatmap to match the target Gaussian heatmap.
All keypoints are treated equally, with the same Gaussian and sigma.
The training focuses only on keypoints within the activation window, meaning the model never encounters an empty heatmap.
Since the input of the model often extends beyond the bounding box to capture more context, the activation window can include areas outside the image.
This enables a limited prediction of keypoints outside the image, although the models are not specifically trained for this task.
For more details, see \cref{sec:keypoints-analysis} in the supplementary.

Heatmap-based approaches can predict keypoints outside of the image to some degree.
Regression methods like RLE \cite{RLE} go even further and predict keypoints further from the image.
Our goal is not to localize out-of-image keypoints per se but emphasize their importance in training and evaluation.

Alternative loss functions have also been proposed.
Adaptive Wing Loss \cite{AdaptiveWingLoss} improves on original Wing Loss \cite{WingLoss}.
CornerNet \cite{CornerNet} introduces a modified Focal Loss (originally in \cite{FocalLoss}) for object detection.
RTMO \cite{RTMO} treats pose estimation as coordinate classification, assuming a normal distribution with varying sigmas for different keypoints, and trains with a negative log likelihood loss.
Distribution-based losses such as \cite{HeatmapDistrMatching} or Kullback-Leibler divergence have also been used for heatmap-based pose estimation.
Most importantly, \cite{YoloPose} introduce a new OKSLoss, that is directly tied to the evaluation metric. Their OKSLoss is computed on the prediced keypoints.
We modify it to use pixel-wise to optimize probability maps without any shape assumptions.

This paper also applies a cropping data augmentation technique.
Similarly to information-dropping methods \cite{InformationDropping} like Hide\&Seek or random bounding box masking, this technique introduces the domain shift by creating more invisible keypoints during training, encouraging the model to leverage the structure of the human body.

The most widely used dataset for 2D human pose estimation is COCO \cite{COCO}, with MPII \cite{MPII} being a less common alternative.
Datasets like OCHuman \cite{OCHuman} and CrowdPose \cite{CrowdPose} focus on multibody problems such as occlusion and self-occlusion.
For COCO and related datasets, the evaluation metric is Object Keypoint Similarity (OKS), while Percentage of Correct Keypoints (PCKh) is used for MPII.

However OKS ignores unannotated keypoints, as COCO is not fully annotated -- many individuals have missing keypoints.
Missing annotations can occur either because the keypoint is outside the image or because of occlusion.
All annotated keypoints in current datasets are within the image, as annotating keypoints outside the image would be too costly and imprecise.
Additionally, OKS does not penalize guessing, and if a model predicts a keypoint inside the image when it is not present, there is no penalty.
Evaluating the presence probability of keypoints (whether they are in the activation window) is crucial for real-world applications and requires out-of-image annotations.

We address these issues by introducing a new dataset, CropCOCO, created by cropping COCO images, along with a new evaluation metric, Extended OKS (Ex-OKS), which accounts for keypoints outside the image.


\section{Method}
\label{sec:method}


ProbPose introduces several key innovations:

\begin{enumerate}
\item A novel approach to keypoint localization using \textit{probability maps}.

\item A different application of OKSLoss \cite{YoloPose} corresponding to expected risk minimization.

\item A new attribute for each keypoint, termed \textit{presence probability}.

\item A data augmentation technique to handle keypoints outside the image.

\item A double-heatmap approach that expands the field of view, enabling localization of keypoints positioned far outside the image.

\end{enumerate}

In combination with visibility and quality estimation, these elements provide a complete description of each keypoint's state in ProbPose -- indicating whether it is within the image, visible, and the confidence level of the estimate.
The probability map offers calibrated localization estimates.
Each component is explained in detail in the following sections.

\subsection{Probability maps and loss function}
\label{sec:method-probmaps}

In conventional methods, heatmaps are trained with Gaussian targets and MSE loss, resulting in fast convergence, but a Gaussian-shaped bias in output.
However, there is no theoretical basis for assuming that the posterior localization probability follows a Gaussian distribution.
The normal distribution in OKS metric reflects human-induced error in the annotation process, but we assume the underlaying distribution is not Gaussian but reflects the body shape.
During inference, heatmaps typically use the argmax for point localization and the peak value as a combined measure of localization quality and visibility.

In our approach, we require the model to output probability maps rather than traditional heatmaps.
Unlike heatmaps, probability maps always sum to 1, achieved through Sparsemax \cite{Sparsemax} as the final activation function.

Each pixel in the probability map represents the posterior probability
\begin{align}
    p_L(x_i) = p(x_i | k_j \in AW, img),
\end{align}
where $x_i$ is pixel coordinate, $k_j$ is the j-th keypoint and \textit{AM} stands for activation window.
With Sparsemax, each probability map sums to 1 for each keypoint:
\begin{align}F
    \sum_{x_i \in AW} p(x_i | k_j \in AW, img) = 1
\end{align}
We refer to this as the \textbf{localization probability} \( p_L(x_i) \).

\subsubsection{Loss function}

We train the probability maps using a modified version of OKSLoss \cite{YoloPose}, aligned with the evaluation function.
The loss is formulated as an expected risk minimization problem.

\begin{align}
    R_{exp}(x_i) &= \big(1 - OKS(x_i)\big) \cdot p_L(x_i) \\
    \mathcal{L}_{OKS}(x_i) &= (1-\alpha) R_{exp}(x_i) + \alpha g(x_i)
    \label{eq:OKS-loss}
\end{align}

In \cref{eq:OKS-loss}, $R_{exp}(x_i)$ represents expected risk and $\mathcal{L}_{OKS}(x_i)$ is the loss function for pixel $x_i$.
In contrast to \cite{YoloPose}, we apply OKSLoss on each pixel of probability map instead of on predicted keypoints.
This loss encourages pixels with low OKS to have low localization probability and those with high OKS to have higher localization probability.

The term $g(x_i)$ is the gradient of the heatmap, computed with the Sobel operator, and serves as a regularizer controlled by hyperparameter $\alpha$.
Without regularization, the probability map quickly forms a sharp peak and overfits to training data.
Smoothing regularization minimizes differences between neighboring pixels, promoting a smooth distribution without assuming a specific shape, as Gaussian regularization does for MSE-trained heatmaps.

\begin{figure}[tb]
  \centering
  \begin{subfigure}{0.495\linewidth}
    \centering
    \includegraphics[width=\linewidth,page=13]{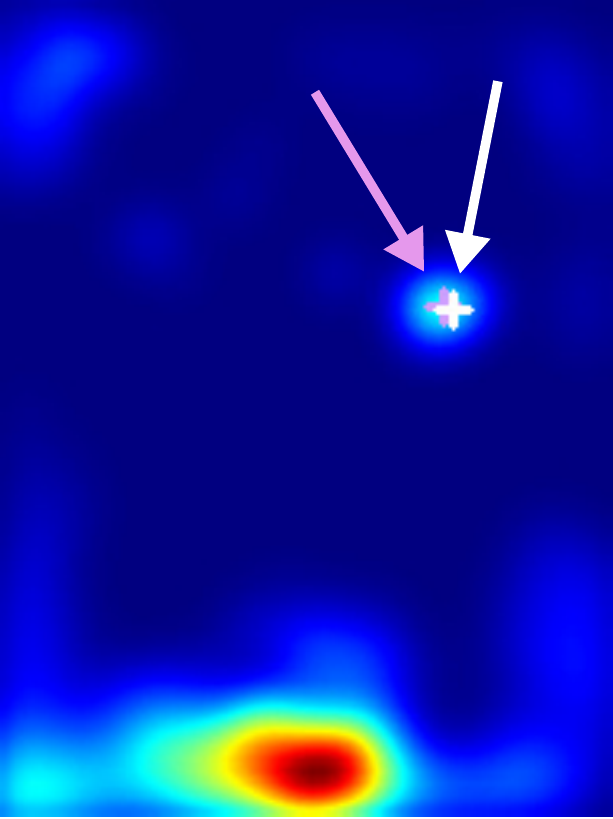}
    \caption{Predicted probability map}
    \label{fig:UDPvsOKS-UDP}
  \end{subfigure}
  \hfill
  \begin{subfigure}{0.495\linewidth}
    \centering
    \includegraphics[width=\linewidth,page=14]{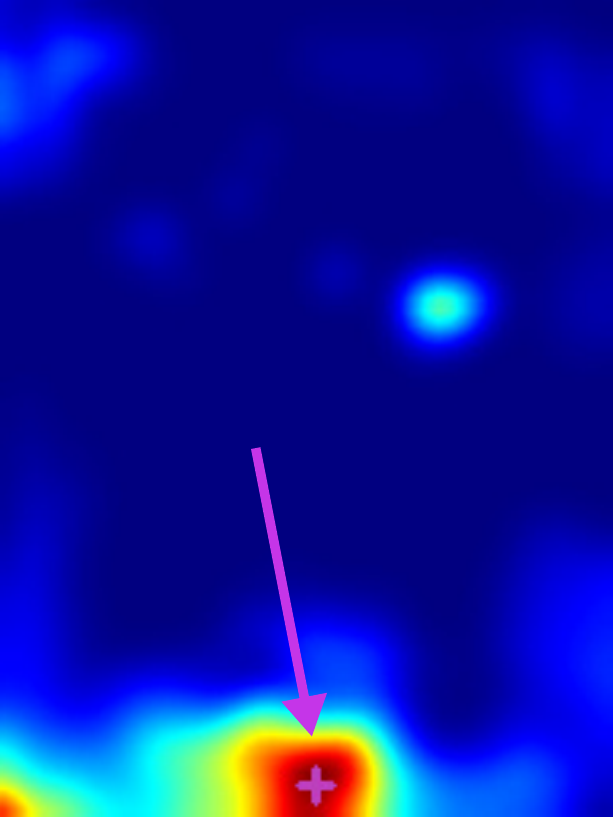}
    \caption{Expected OKS}
    \label{fig:UDPvsOKS-OKS}
  \end{subfigure}
  \caption{
  UDP decoding vs. expected OKS maximization.
  In the left image, white cross is maximum value, light purple is maximum refined by UDP.
  In the right image, the dark purple cross is maximum expected OKS.
  Maximizing expected OKS is robust to sharp local peaks and prefer areas with big mass compared to simple maxima.
  }
  \label{fig:UDPvsExpOKSMax}
\end{figure}

For localization, instead of directly using argmax, we compute the expected OKS for each pixel and take its argmax, refining it with quadratic interpolation for sub-pixel precision.
This approach, based on expected OKS rather than UDP decoding, slightly enhances localization, especially in cases of bimodal heatmaps like in \cref{fig:UDPvsExpOKSMax}.

Probability maps offer improved versatility.
They allow point estimates, as required by the COCO evaluation protocol, and enable probabilistic queries, such as defining the smallest area containing a keypoint with at least 90\% probability.
However, the peak of the probability map can no longer reliably indicate the quality of the localization or the presence of keypoints in the activation window.
The former is addressed by predicting the OKS for each keypoint as in \cite{Calibration}, and the latter is addressed by directly predicting the keypoint's \textit{presence probability}.

\subsection{Presence probability}
\label{sec:method-presence}

Previous models and evaluation protocols have overlooked the question of whether a keypoint is within the activation window.
Top-down approaches assume that the bounding boxes fully enclose all keypoints, but this is not always the case due to factors such as occlusion or crop (see \cref{sec:keypoints-analysis} in the supplementary for further discussion).
Thus, it is essential for the model to assess whether a keypoint is actually present in the activation window.
This is usually addressed by thresholding confidence (the maximum heatmap value) with an arbitrary cut-off.
However, evaluation protocols have ignored keypoint presence, with models predicting a location regardless.

\textbf{Presence probability} $p(k_j \in AW | img)$ complements probability maps, which inherently assume that keypoints are within the activation window.
If the keypoint is outside, the presence probability reflects this; if inside, the probability map provides the location.
Visual explanation is in \cref{fig:am-visualization}.
We abbreviate the \textit{presence probability} to $p_p(k_j)$  and train it with Binary Cross Entropy loss.

To train the presence probability, we need keypoints that are certainly outside the activation window.
These samples are generated through various data augmentation techniques, such as scaling and rotation, with \textit{random cropping} as the primary source.

\subsection{Calibration}

Both the probability maps and presence probability are calibrated.
Probability maps are calibrated so that the top 5\% of the map contains 5\% of the keypoints, the top 10\% contains 10\% of the keypoints, and so on.
We achieve this calibration through temperature scaling on CropCOCO.
As a result, the calibrated probabilities align with the true underlying distribution, enabling more versatile evaluations beyond simple point estimates.
Details and calibration curves are in \cref{sec:model-calibration} in the supplementary.

\subsection{Crop image augmentaton}
\label{sec:method-crop}

As discussed in \cref{sec:keypoints-analysis}, we consider the image border as a type of occlusion.
To train the model to correctly identify keypoints outside the image, we need suitable training samples.
Manual annotation of such keypoints is both costly and imprecise, so we generate these samples by cropping existing annotated keypoints out of the image.

Depending on the crop strength, the in-the-image keypoint may end up within the activation window but outside the image, or entirely outside the activation window.
Keypoints within the activation window but outside the image are used to train probability maps, while those outside the activation window train the presence probability.

The cropping not only provides data for presence probability training but also enhances the model’s ability to localize keypoints near the image boundary, extending even to areas beyond the visible border.
This is especially useful for cases where the subject is partially cropped by the image border or in close-up pose estimation, where keypoints may lie far from the visible body.

Examples of crops and results visualizations are in the \cref{fig:OoI-examples,fig:ooi-comparison}.

\subsection{Enlarging activation window size}
\label{sec:method-multi}

When predicting keypoints outside the image, a key question is how far from the image we aim to localize them.
In applications where out-of-image keypoint localization is essential, it might be helpful to enlarge the activation window.
However, simply increasing the input size or activation window typically reduces performance.

\begin{figure}[t]
  \centering
  \begin{subfigure}{0.495\linewidth}
    \centering
    \includegraphics[width=\linewidth,page=4]{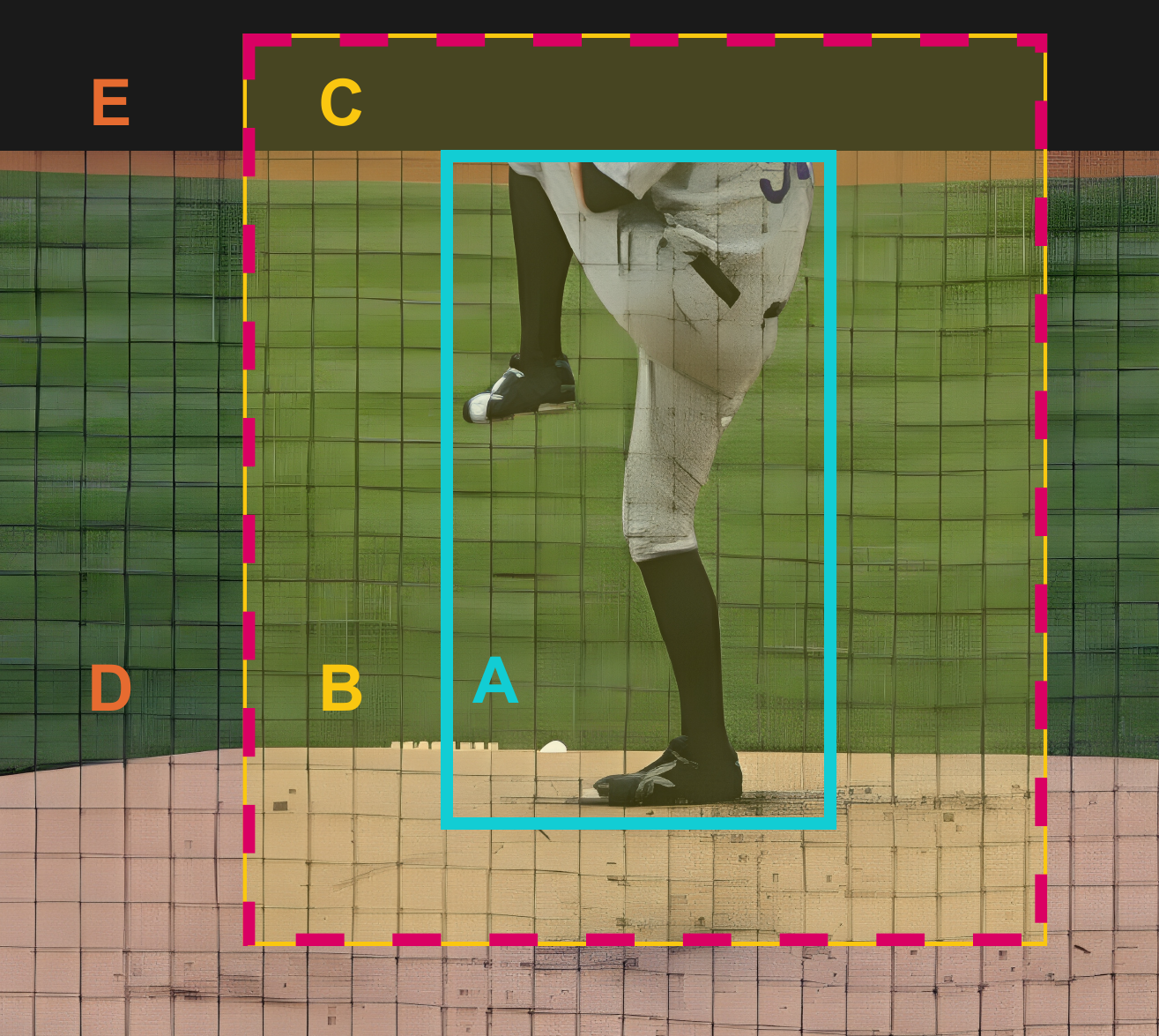}
    \caption{Traditional top-down method}
    \label{fig:am-visualization-single}
  \end{subfigure}
  \hfill
  \begin{subfigure}{0.475\linewidth}
    \centering
    \includegraphics[width=\linewidth,page=3]{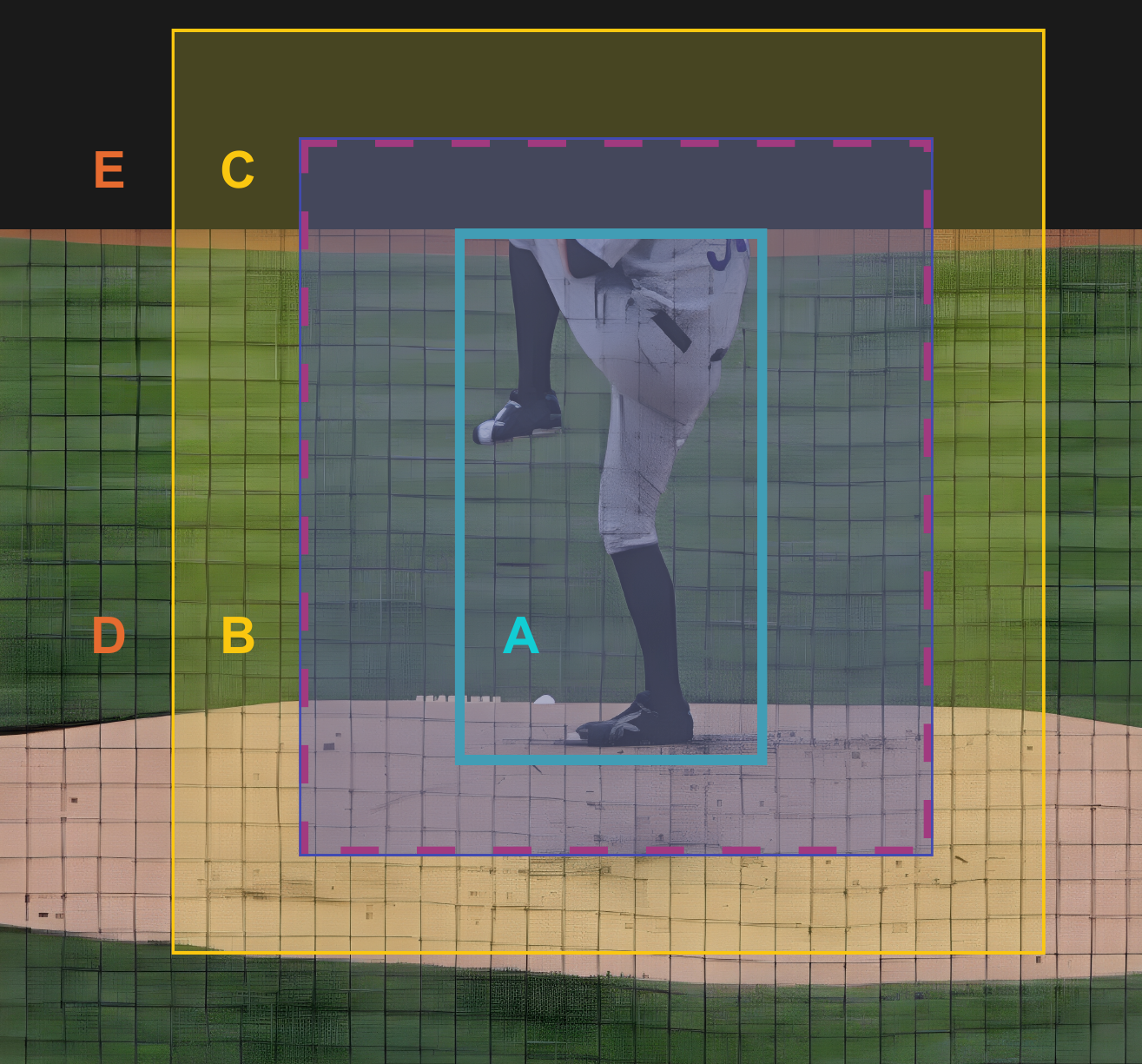}
    \caption{Double-heatmap approach}
    \label{fig:am-visualization-double}
  \end{subfigure}
  \caption{
  Visualizations of top-down approaches.
  Rectangles represent \colorbox{cyan}{bounding box},
  \colorbox{magenta}{model input} and
  \colorbox{yellow}{activation window}.
  Right image shows double-heatmap approach with second \colorbox{blue}{\textcolor{white}{activation window}}.
  Probability map is responsible for localization of keypoints in the activation window (A, B, C) and presence probability predict if the keypoint is outside (D, E).
  Note that the double-heatmap approach has larger field-of-view.
  For detailed discussion about keypoint location, see \cref{sec:border-points} in supplementary.
  }
  \label{fig:am-visualization}
\end{figure}

To address this, we propose a double heatmap approach (illustrated in \cref{fig:am-visualization-double}).
In this method, an expert heatmap precisely localizes keypoints within a smaller activation window that matches the model input, while a larger activation window captures keypoints further from the image.
The larger probability map has the same resolution but a wider field of view, resulting in a lower effective resolution.
Both maps are trained on keypoints within their respective activation areas.
If the larger map predicts a keypoint within the smaller activation area, the expert map refines the localization for greater accuracy.

This approach balances the trade-off between field-of-view and precision.
The larger map offers a broader field of view, but with lower accuracy, while the expert map maintains precision within a smaller activation window.
This method is particularly suitable for applications like close-up pose estimation, where a wide field of view is essential.

\section{Data}
\label{sec:data}

To evaluate ProbPose, we introduce a new evaluation protocol based on OKS and mAP, along with a new dataset specifically designed for out-of-image keypoint localization and presence probability classification.

\subsection{Evaluation metrics}
\label{sec:data-eval}

Current metrics, such as OKS and PCK, evaluate only in-image keypoints, overlooking false positives.
As a result, models overestimate confidence, guess keypoint locations, and are not penalized for predicting absent keypoints.

We propose a new evaluation approach that checks whether a keypoint is within the activation window and, if so, assesses its precise localization.
This approach better reflects real-world scenarios, like human-machine interaction, where knowing if a keypoint cannot be localized is critical.
To address this, we introduce Extended OKS (Ex-OKS), an extension of OKS that accommodates keypoints outside the activation window.
For bottom-up and single-stage methods, where the activation window spans the entire image, Ex-OKS can be adapted accordingly.

In its localization-only form, Ex-OKS functions identically to the original OKS.
However, Ex-OKS extends OKS by accounting for keypoints outside the activation window.
The essential goal is for the model to indicate if a keypoint is absent, rather than attempting a guess.

Ex-OKS evaluates two variables for each keypoint: location $x_i$ and presence probability $p_p$.
Although the model outputs a continuous value for $p_p$, a binary decision (0 or 1) is often required in real-world applications, so we optimally threshold it for each model.

Formally, we define \textbf{Ex-OKS} as:

\begin{align}
    &d_i = \begin{cases}
        d_e(x^{*}_i, x'_i) & \text{if } p^{*}_p = 1 \text{ and } p'_p = 1 \\
        d_e(\text{AM}, x'_i) & \text{if } p^{*}_p = 0 \text{ and } p'_p = 1 \\
        d_e(x^{*}_i, \text{AM}) & \text{if } p^{*}_p = 1 \text{ and } p'_p = 0 \\
        0                               & \text{else} 
    \end{cases} \\
    &\text{Ex-OKS} = \exp{(\frac{-d_{i}^2}{2k^2\sigma^2})}
    \label{eq:Ex-OKS}
\end{align}
, where \textit{AM} stands for activation window, $d_e$ is Euclidean distance and $\cdot^*$ and $\cdot'$ are ground truth and predicted variables, respectivelly.

If both the ground truth and the model agree that $p_p = 1$, Ex-OKS defaults to the standard OKS evaluation.

When the ground truth $p^{*}_p = 0$ and the model predicts $p'_p = 1$, the penalty depends on the predicted location $x'_i$ and its distance from the edge of the activation window, with lower penalties for points near the boundary.
Similarly, if $p^{*}_p = 1$ but the model predicts $p'_p = 0$, the penalty is based on the ground truth location $x^{*}_i$, with smaller penalties for errors near the border of the activation window.

When both $p^{*}_p = 0$ and $p'_p = 0$, there is no penalty and the similarity score is 1.

Ex-mAP, built on Ex-OKS, extends mAP by incorporating both keypoint localization and presence probability, making it suitable for evaluating models with keypoints outside the activation window.

\subsection{Datasets}
\label{sec:data-datasets}

We evaluated our model on the COCO dataset to ensure a fair comparison with existing models. All training was conducted on COCO, with crop image augmentation applied in specified experiments.

To assess performance on out-of-image keypoints, we introduce a new dataset, CropCOCO.
CropCOCO consists of randomly cropped COCO validation images, with some keypoints positioned outside the activation window.
Bounding boxes are recomputed for the cropped images to evaluate the model appropriately.
While other researchers can create their own cropped dataset, we release CropCOCO to support reproducibility.

During COCO evaluation, we observed that keypoints near the image boundary are often annotated further inside the image, likely due to annotators avoiding placements near the edge.
This causes our model to be penalized for accurately localizing keypoints outside the image.
When keypoints near the border are excluded from evaluation, ProbPose demonstrates even stronger performance compared to other state-of-the-art methods.
Although this issue is minor (causing about a 0.2\% performance difference) and difficult to correct, we highlight it to raise awareness of its potential impact.
For further details, see \cref{sec:border-points}.

Additionally, we evaluated our model on the OCHuman dataset to demonstrate its generalizability across different domains.

\section{Experiments}
\label{sec:experiments}

\begin{table*}[tb]
    \centering
    \begin{tabular}{@{}l||rr|rr|rr@{}l}
        \toprule
        \multirow{2}{*}{Model} & \multicolumn{2}{c|}{COCO}  & \multicolumn{2}{c|}{CropCOCO} & \multicolumn{2}{c}{OCHuman}  \\
                        & mAP  & Ex-mAP & mAP & Ex-mAP & mAP  & Ex-mAP  \\
        \midrule
        HRFormer-s \cite{HRFormer}      & 75.2 & 74.6   & 70.9 & 64.3  & 60.3 & 60.0    \\
        PVTv2 \cite{PVTv2}           & 72.0 & 71.5   & 70.7 & 63.1  & 58.5 & 58.1    \\
        SWIN-t \cite{SWIN}          & 73.5 & 72.9   & 71.3 & 65.0  & 58.1 & 57.9    \\
        ViTPose-s \cite{ViTPose}       & 75.9 & 75.3 & 72.7 & 66.5  & 60.3 & 60.1 \\
        ProbPose-s      & \textbf{76.6} & \textbf{76.4} & \textbf{81.7} & \textbf{73.9} & \underline{60.4} & \underline{60.2} \\
        ProbPose-s-DH             & \underline{76.2} & \underline{75.4} & \underline{80.9} & \underline{71.4}  & \textbf{61.4} & \textbf{61.2} \\
        \bottomrule
    \end{tabular}
    \caption{
      Keypoint localization accuracy in terms the of the standard mAP and the novel Ex-mAP metrics. 
       ProbPose is compared with SOTA models of similar size on the COCO, CropCOCO, and OCHuman datasets.
        All evaluation is done using GT bounding boxes. DH stands for the double heatmap approach with a wider field of view.
        To compute Ex-mAP, we select the optimal threshold for each model.
        The best model is in \textbf{bold}, the second-best  is \underline{underlined}. ProbPose brings significant improvement for out-of-image keypoints and minor improvement on standard COCO.
    }
    \label{tab:SOTA-comparison}
\end{table*}

The results in \cref{tab:SOTA-comparison} compare our method with other SOTA approaches on COCO, CropCOCO, and OCHuman datasets using mAP and Ex-mAP metrics.
For fair comparison, we evaluate models of similar size, with some using the same backbone (ViT-s).
All experiments use ground truth bounding boxes.

ProbPose improves on its baseline model, ViTPose, by approximately 1\% in localization, a gain comparable to that achieved by other information-dropping augmentations like \cite{InformationDropping}.
Importantly, ProbPose demonstrates stronger performance in Ex-OKS, showing improved detection of keypoints outside the activation window due to its direct \textit{presence probability} prediction, as discussed in \cref{sec:presence-probability-results}.
Since COCO lacks out-of-image keypoints, Ex-OKS improvement here reflects enhanced presence probability classification.
Localization metrics could be slightly higher if not for incorrect annotations near bounding box borders, discussed further in \cref{sec:border-points}.

The most significant improvement is unsurprisingly observed on CropCOCO, where ProbPose excels at detecting out-of-image keypoints, primarily due to crop data augmentation.
Again, the localization metrics could be slightly higher without annotation errors along bounding box borders.

ProbPose-DH, the double-heatmap variant, shows a slight performance decrease on COCO and CropCOCO, reflecting the trade-off between a broader field of view and in-image precision (see \cref{sec:method-multi}).
In this variant, each heatmap is predicted by one head from shared features, and the larger heatmap decides when to defer to the more precise heatmap.
If the larger heatmap estimates the wrong localization, the specialized heatmap cannot correct it.
Additionally, the presence probability boundary in ProbPose-DH is beyond the image border, requiring an expanded field of view, which lowers the effective heatmap resolution and impacts accuracy.
However, the decrease is minor, and ProbPose-DH remains competitive with other SOTA methods, providing a better alternative to simply enlarging the input size or activation window (which reduces resolution).
Interestingly, ProbPose-DH shows a notable improvement in OCHuman, likely due to its extended field of view, which enhances individual differentiation in complex scenes.

Both ProbPose variants perform well on OCHuman, highlighting their strong generalization across domains.
This suggests ProbPose’s value in real-world applications, as it adapts effectively to datasets unseen during the training.

Overall, ProbPose matches the in-image localization improvements of other information-dropping models while significantly enhancing out-of-image keypoint detection, robustness along bounding box edges, and model versatility.
Its calibrated presence probability also makes ProbPose particularly suitable for real-world applications.

\subsection{Presence probability vs. confidence}
\label{sec:presence-probability-results}

\begin{figure}[tb]
    \centering
    \includegraphics[width=\linewidth]{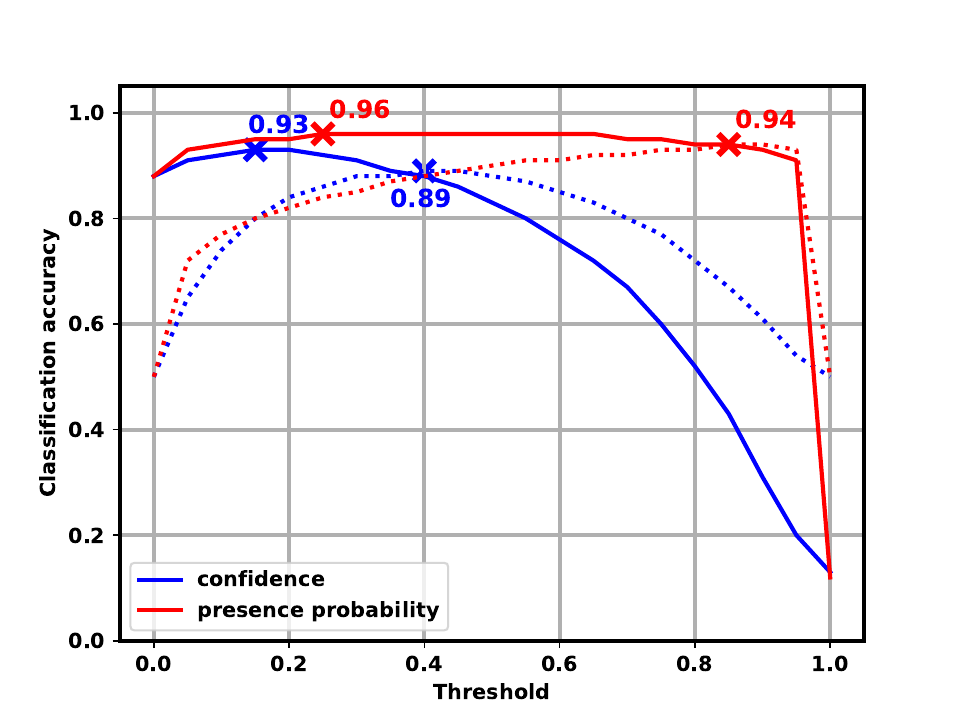}
    \caption{ Keypoint presence in the activation window.
    Comparison of accuracy of the in vs. outside the activation window classification based on thresholding  of the ProbPose's presence probability and confidence of the ViTPose model.
    The accuracy is measured for CropCOCO dataset (full line) and its balanced subset (dotted line).
    The proposed Presence probability outperforms ViTPose confidence in both cases, reducing the error by 30\% and 45\% respectively.}
    \label{fig:presence-probability}
\end{figure}

\cref{tab:SOTA-comparison} highlights the advantages of using presence probability over confidence, reflected in the increase in Ex-mAP.
To clarify these results, we provide an illustrative example in \cref{fig:presence-probability}.

\cref{fig:presence-probability} shows classification accuracy for CropCOCO and its balanced subset.
To get the balanced subset, we randomly selected keypoints from both classes to have the same amount of each.
The task is to classify whether keypoints are inside or outside the activation window, indicating if they should be drawn or used in other applications.
In both balanced and unbalanced datasets, the presence probability outperforms the confidence score from the ViTPose model.
This improvement is due to confidence not being trained to reflect presence probability, and the standard COCO dataset, without crop data augmentation, lacks sufficient training samples for this purpose.

\cref{fig:presence-probability} also provides insights into confidence thresholding.
Current models often use an arbitrary threshold of 0.3 to decide whether to trust predictions without theoretical justification.
The evaluation on CropCOCO reveals that the optimal confidence threshold is between 0.15 (for unbalanced datasets) and 0.4 (for balanced datasets).
Thus, the commonly used threshold of 0.3 is close to the optimum, which aligns with its generally good results.
This study offers guidance on setting an appropriate threshold based on the characteristics of a specific dataset.

\subsection{Ablation study}
\label{sec:experiments-ablation}

\begin{table}[tb]
    \centering
    \begin{tabular}{@{}ll||rr@{}l}
        \toprule
        crop & p-maps & COCO & CropCOCO  \\
        \midrule
        \xmark & \xmark &  76.0  & 73.7 \\   
        \xmark & \cmark &  76.4  & 72.4 \\   
        \cmark & \xmark &  76.6  & 81.7  \\    
        \cmark & \cmark &  76.6  & 81.7  \\       
        \bottomrule
    \end{tabular}
    \caption{PropPose ablation; mAP on COCO and CropCOCO for versions without the probability maps and crop data augmentation. Crop augmentation clearly helps for cropped images. Probability maps bring negligible performance improvement as their main strength is in versatility.}
    \label{tab:ablation-study}
\end{table}

\cref{tab:ablation-study} presents an ablation study evaluating the core components of our approach, probability maps and crop data augmentation.
The analysis is conducted on the ProbPose-s model using ground truth bounding boxes on the COCO and CropCOCO datasets.

Using probability maps without crop data augmentation yields the lowest performance.
This setup slightly improves localization on COCO but reduces the accuracy of out-of-image keypoints in CropCOCO.
Without crop data augmentation, ProbPose encounters few out-of-image training samples, and probability map normalization forces predictions within the activation window, making probability maps less effective than heatmaps for out-of-image keypoints.

Adding crop data augmentation improves localization accuracy on both COCO and CropCOCO.
Training with out-of-image keypoints also improves in-image localization, as the model relies more on the structure of the human body, similar to \cite{InformationDropping}.
Compared to Hide\&Seek, crop augmentation brings the same in-image performance improvement, but also improve out-of-image localization and generate samples for training \textit{presence probability}.

Combining crop augmentation and probability maps achieves the same localization accuracy as using crop data augmentation alone.
However, as discussed in \cref{tab:SOTA-comparison}, the probabilistic approach is more important for presence probability classification.
While COCO-like evaluation protocols focus on point estimates, probability maps are more versatile and provide direct mathematical interpretation.
Heatmaps are optimized for point estimates, as in COCO evaluations but probability maps support broader applications without compromising point-estimate accuracy.



\section{Conclusions}
\label{sec:conclusions}

We present ProbPose, a novel approach to 2D human pose estimation that uses probability maps instead of heatmaps and introduces presence within-the-activation-map probability prediction.
Unlike previous methods, we do not assume a Gaussian distribution for the localization probability and employ the expected OKS for both training and decoding, thus aligning the training loss and the evaluation metric. 
The key findings are:

\begin{enumerate}
    \item Probability maps with OKSLoss improve versatility without sacrificing performance, offering more flexibility than point estimates.

    \item Modified OKSLoss adapts to each keypoint type, enforcing sharper distributions for high-precision keypoints like eyes or nose, while broader distributions are used for less precise keypoints like hips. These distributions are more reflective of the image and are not constrained by a Gaussian assumption.

    \item Predicting presence probability is more effective than thresholding confidence, reducing error by 45\%. Presence probability is a better indicator for in/out decisions than the peak of a predicted Gaussian.

    \item Cropping images during training improves both out-of-image localization and in-image performance by increasing data variability, similar to other information-dropping augmentations. Unlike Hide\&Seek \cite{InformationDropping}, crop augmentation also enables out-of-image localization.

    \item Increasing the activation window size comes with some performance trade-offs, but the impact can be minimized. Our double-heatmap model can localize keypoints up to 25\% outside the bounding box with only an 0.4\% decrease in AP performance.
    
\end{enumerate}

Future work could explore the behavior of probability maps in multibody scenarios.
Decoding with expected OKS maximization has proven especially useful for bimodal heatmaps, which are common in multibody images.

The issue of COCO keypoints near the edge of the bounding box remains unresolved.
Ignoring these keypoints during training reduces performance as valuable data is lost.
However, training with only COCO keypoints teaches the model to place keypoints further inside the image.
Cropping as a data augmentation strategy helps mitigate this issue during training but not during evaluation.
Re-annotating the dataset would be too costly and likely imprecise, as it is difficult for humans to accurately place keypoints outside the image.

{\small
\bibliographystyle{ieee_fullname}
\bibliography{main}
}

\appendix
\maketitlesupplementary

\section{On the types of keypoint}
\label{sec:keypoints-analysis}

\begin{figure}
    \centering
    \includegraphics[width=\linewidth,page=2]{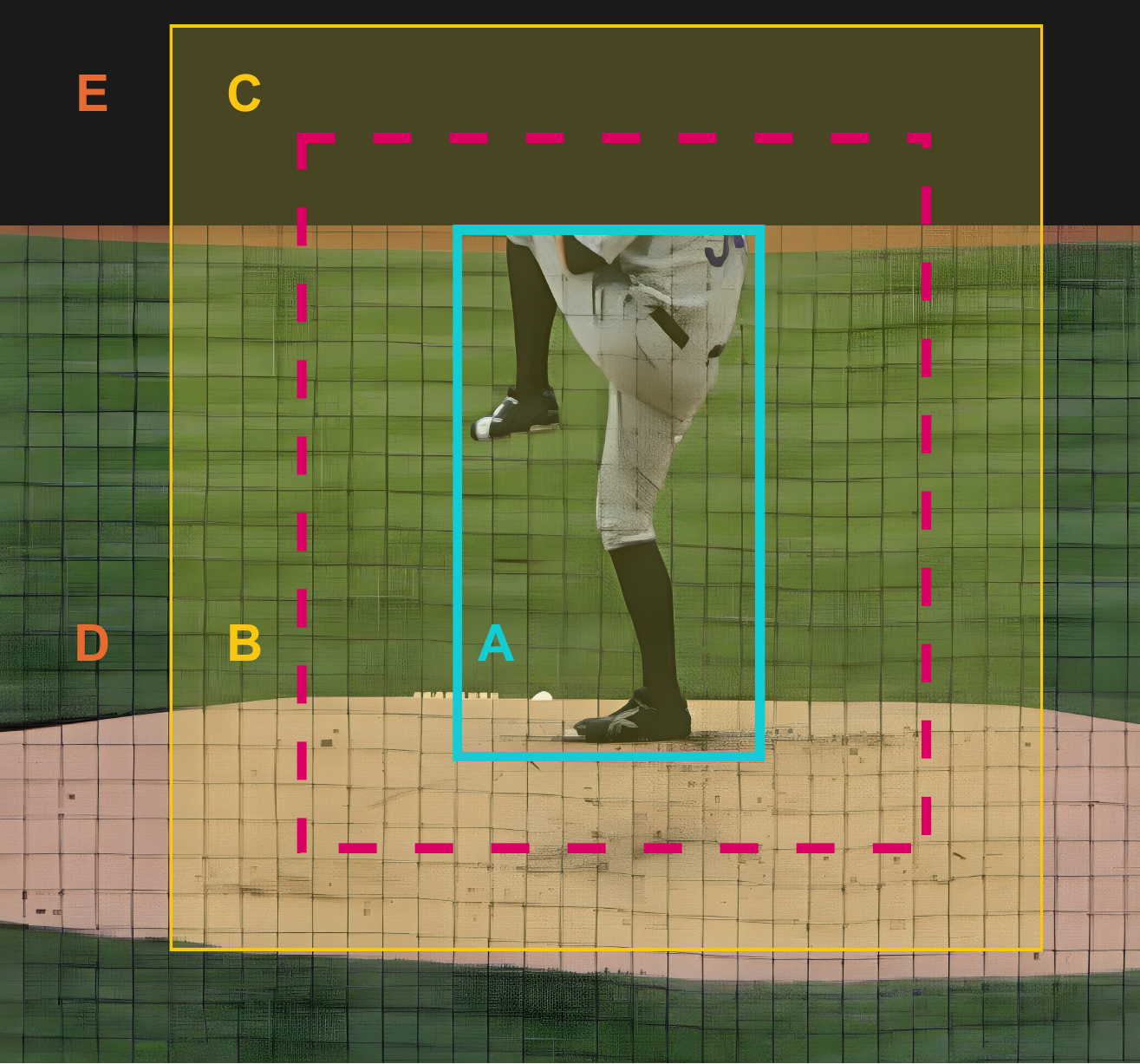}
    \caption{
    A scheme explaining where keypoints could be in the image.
    Rectangles represent the \colorbox{cyan}{bounding box}, the \colorbox{magenta}{model input} and the \colorbox{yellow}{activation window} (usually coincides with model input).
    Image taken from the COCO val dataset.
    }
    \label{fig:bbox-scheme}
\end{figure}

\cref{fig:bbox-scheme} illustrates possible keypoint locations, assuming that all keypoints of an individual are present within or outside the image.
The image edge is treated as another form of occlusion, much like an object blocking part of the body.
Whether a person is occluded by an object (e.g., a wardrobe) or cropped by the image, invisible keypoints must be estimated from the visible ones and the structure of the human body.

In the top-down approach to human pose estimation, the image is divided into three main areas, which are depicted in \cref{fig:bbox-scheme} and defined below.

\textbf{\colorbox{cyan}{Bounding box}} (bbox) -- the tightest rectangle enclosing all visible parts of the individual. A perfect human detector outputs this kind of bounding box.

\textbf{\colorbox{magenta}{Model input}} -- the part of the image cropped and fed into the top-down model. Due to aspect ratio constraints and the need for contextual information, the model input is usually larger than the bounding box and often includes areas outside the image.

\textbf{\colorbox{yellow}{Activation window}} (amap, AM) --  the area where the top-down model localizes keypoints. This typically coincides with the model input but can be larger or smaller. Like the model input, the activation window often contains regions outside the image.

The bounding box, activation window, and image edge divide the space into five subareas, each behaving differently in the context of top-down human pose estimation. These subareas (A-E) are visualized in the \cref{fig:bbox-scheme}.

\begin{enumerate}[label=\Alph*]
    \item -- \colorbox{LightCyan}{inside the bbox}. Visible keypoints can only exist within the bounding box.
    
    \item -- \colorbox{LightYellow}{inside both the activation window and the image}. The vast majority of COCO keypoints fall into areas A and B. No visible keypoints are located outside the bounding box.
    
    \item -- \colorbox{LightYellow}{inside the activation window but outside of the image}. Previous methods could theoretically predict keypoints in area C, but they lack the necessary training data to do so.
    
    \item -- \colorbox{LightOrange}{outside of the activation window but inside the image}. Prior top-down methods cannot localize keypoints in this area or describe them in any way. Approximately 0.2\% of keypoints in the COCO dataset fall into this category, meaning top-down methods are always penalized by OKS for these points. However, ProbPose marks these keypoints as "out" by predicting low presence probability and won't get penalized by Ex-OKS.
    
    \item -- \colorbox{LightOrange}{outside of both the image and the activation window}. Like points in area D, keypoints in area E have been ignored by previous methods in both estimation and evaluation. ProbPose, along with Ex-OKS, addresses this issue using presence probability and a novel evaluation metric Ex-OKS.
\end{enumerate}

\begin{table}[tb]
    \centering
    \begin{tabular}{l|A B B D D}
        Dataset & A & B & C & D & E \\
        \hline
        COCO train     & 96.2 & 3.5 & 0.0 & 0.2 & 0.0 \\
        COCO val       & 95.8 & 3.9 & 0.0 & 0.2 & 0.0 \\
        CropCOCO val   & 68.8 & 2.2 & 23.5 & 0.1 & 5.3 \\
        OCHuman        & 99.2 & 0.8 & 0.0 & 0.0 & 0.0
    \end{tabular}
    \caption{
    Domain shift between used datasets.
    Percenatages of keypoint types.
    For definitions, see text.
    } 
    \label{tab:domain-shift}
\end{table}

The proportion of annotated keypoints in each area defines the domain of a dataset.
For example, the domain of the COCO-val dataset is represented by the vector (95.8, 3.9, 0, 0.2, 0), where each value indicates the percentage of points in the corresponding subarea.
In particular, there are no annotated keypoints outside the image, and approximately 99.8\% of the keypoints are within the activation window.
Traditional top-down methods assume that 100\% of keypoints lie within the activation window.

\cref{tab:domain-shift} compares the domains of the datasets used for the ProbPose evaluation.
Before this paper, no dataset included annotations outside the image, specifically in areas C and D.
Therefore, no evaluation protocol worked with these areas.
Thus, previous evaluation protocols did not account for these areas.
Area D becomes critical under heavy occlusion, where the detected bounding box is much smaller than the individual.
Likewise, areas C and E become important when the image is heavily cropped or in close-view pose estimation.
The CropCOCO dataset tests the model under domain shift, where keypoints were moved from area A to areas C and E.

\textbf{The visibility} of the keypoint is only loosely related to areas A-E.
Although visible keypoints are always within the bounding box (area A), invisible keypoints can be located in any of the areas.
Importantly, classifying keypoint visibility is a different task from determining whether a keypoint is present in the activation window.

\section{Model calibration}
\label{sec:model-calibration}

Training probability maps with OKSLoss results in uncalibrated probability maps.
To achieve calibrated probability maps, we quantize the probabilities in 5\% increments and ensure each range contains approximately 5\% ground truth keypoints of the COCO validation dataset.
This approach enables probabilistic statements, such as estimating the area where the keypoint is located with 95\% confidence.

\begin{figure}
    
    \begin{subfigure}{\linewidth}
        \centering
        \includegraphics[width=0.495\linewidth]{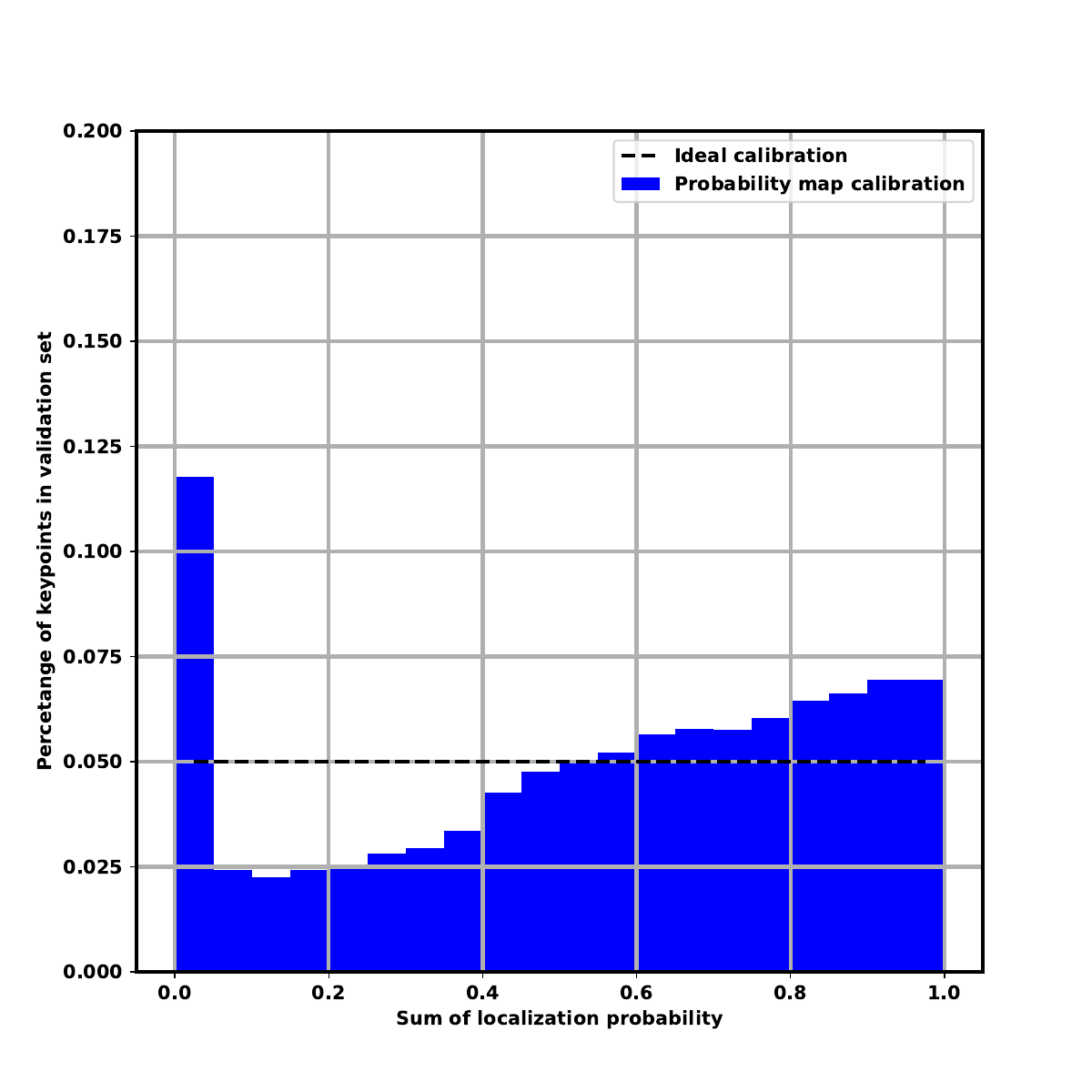}
        \hfill
        \includegraphics[width=0.495\linewidth]{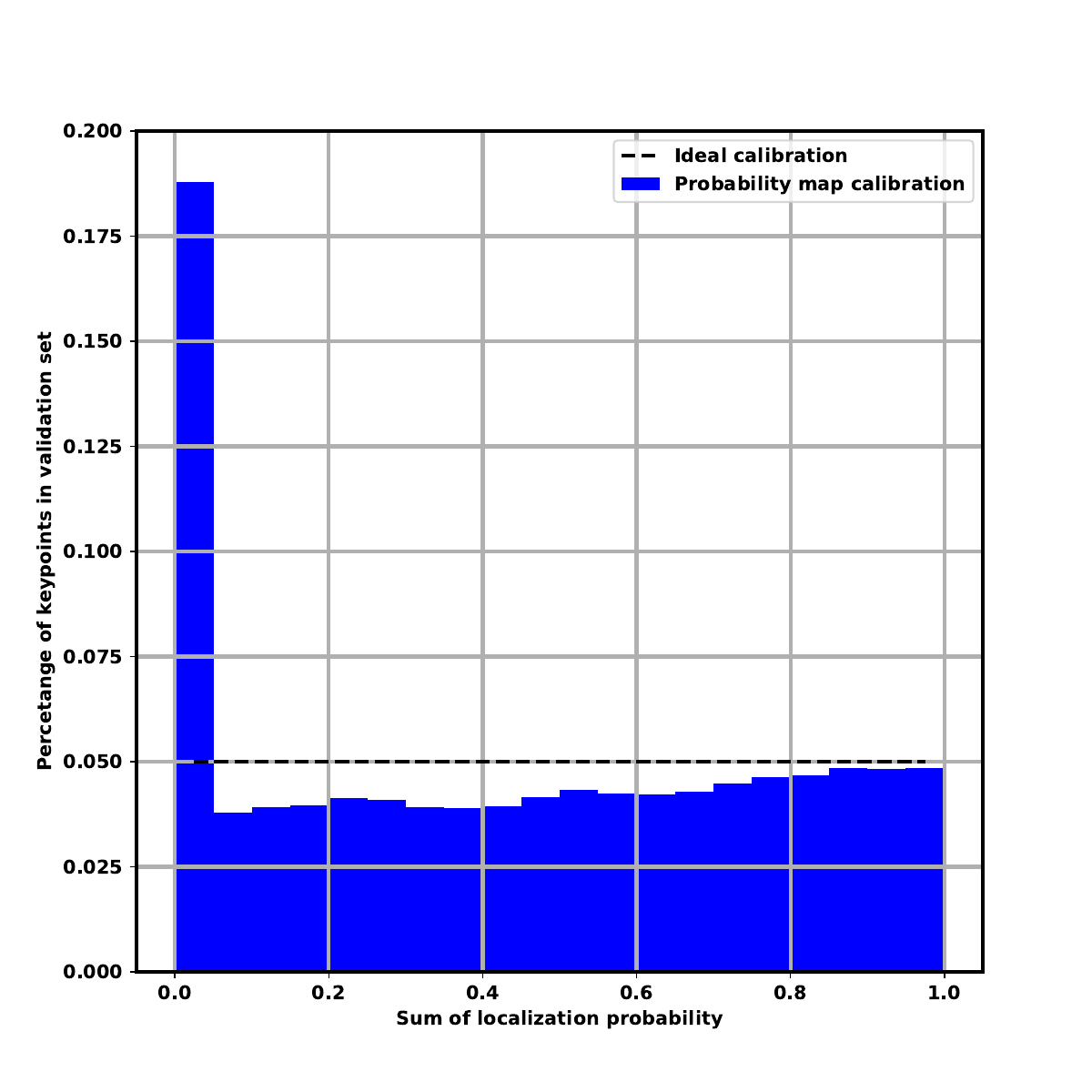}
        \caption{Probability map calibration curves}
        \label{fig:heatmap-callibration}
    \end{subfigure}

    \begin{subfigure}{\linewidth}
        \centering
        \includegraphics[width=0.6\linewidth]{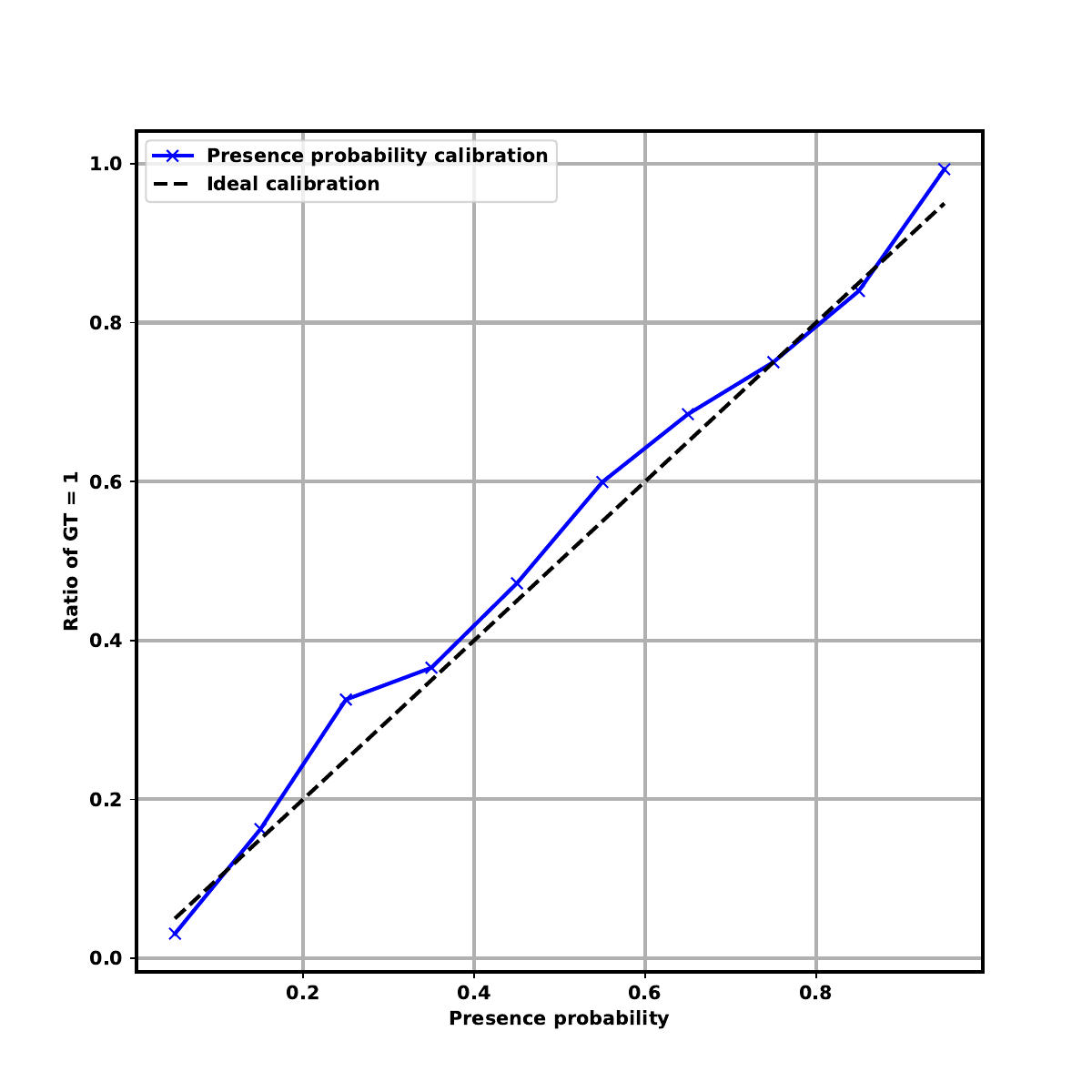}
        \caption{Presence probability calibration curves}\
        \label{fig:presence-probability-calibration}
    \end{subfigure}

    \centering
    \caption{
    Calibration curves of probability maps (a) before and after temperature scaling and presence probability (b).
    While presence probability is calibrated in a standard way, for probability maps we require that top 5\% of pixels contains 5\% of points etc.
    The peak in the calibration of the probability maps between 0\% and 5\% are hard errors where the model fails to locate the keypoint.
    These are usually caused by occlusion and multibody situations.}
\end{figure}

The \cref{fig:heatmap-callibration} shows the calibration curves before and after the temperature scaling.
Notice the large peak in the 0\% to 5\% range.
These correspond to keypoints where the model failed completely, predicting very low probability for the correct area, or incorrectly detecting the keypoint elsewhere ("keypoint stealing" in overlapping individuals).
More research on overlapping individuals or data augmentation techniques such as \cite{AdversarialAugmentation} can help address this issue.

The \cref{fig:presence-probability-calibration} presents the calibration curve of the presence probability.
Unlike keypoint localization, presence probability is naturally calibrated during training due to the use of binary cross-entropy loss and the similar distribution between the training and testing sets.

\section{Expected OKS maximization vs. UDP}
\label{sec:ExpectedOKS}

\textbf{UDP decoding} \cite{UDP} first estimates the global maximum of the predicted heatmap and then refines it to subpixel precision. To refine the localization, the heatmap is blurred using a Gaussian with fixed variance, and the maximum value is shifted toward the estimated peak of the local Gaussian. However, if the initial estimate (heatmap maximum) is incorrect, UDP refining cannot correct it. UDP decoding assumes the predicted heatmap follows a Gaussian distribution and estimates the peak of the predicted Gaussian.

\textbf{Expected OKS maximization}, on the other hand, convolves the predicted probability map with an OKS kernel and calculates the expected OKS for each pixel. The OKS kernel varies for each keypoint type. The initial estimate is the global maximum of the expected OKS map. To achieve subpixel precision, we fine-tune the estimate using quadratic interpolation in the neighboring pixels. Expected OKS decoding makes no assumptions about the shape of the distribution and takes a global approach, favoring areas with larger mass over sharp peaks, aligning more closely with probability maps.

\begin{figure}[tb]
  \centering
  \begin{subfigure}{\linewidth}
    \centering
    \includegraphics[width=0.495\linewidth]{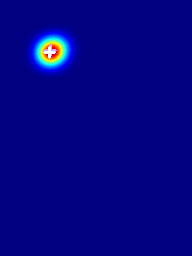}
    \hfill
    \includegraphics[width=0.495\linewidth]{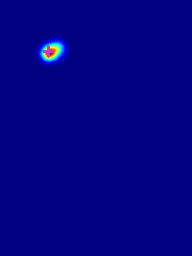}
  \end{subfigure}
  
  \begin{subfigure}{\linewidth}
    \centering
    \includegraphics[width=0.495\linewidth]{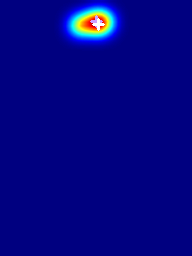}
    \hfill
    \includegraphics[width=0.495\linewidth]{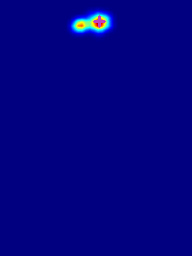}
  \end{subfigure}
  
  \begin{subfigure}{\linewidth}
    \centering
    \includegraphics[width=0.495\linewidth]{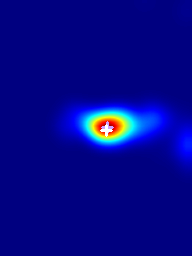}
    \hfill
    \includegraphics[width=0.495\linewidth]{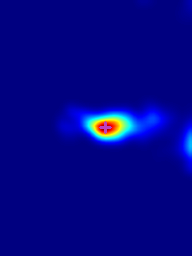}
  \end{subfigure}

    \caption{
    Decoding predicted probability maps throught UDP (left) and expected OKS maximization (right).
    Probability map maximum in white, UDP-refined point in light purple and maximal expected OKS in dark purple.
    These are examples where the decoding does not matter as the predicted heatmap is unimodal.
    Majority of keypoints have such heatmaps so the difference in performance is not big.
    Notice the non-Gaussian shape of predicted probability maps and sharper peaks for expected OKS as opposed to UDP.
    }
    \label{fig:UDP-OKS-good}
  
\end{figure}

\begin{figure}[tb]
  \centering
  \begin{subfigure}{\linewidth}
    \centering
    \includegraphics[width=0.495\linewidth]{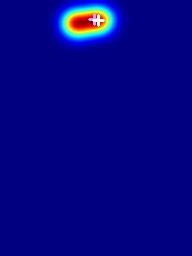}
    \hfill
    \includegraphics[width=0.495\linewidth]{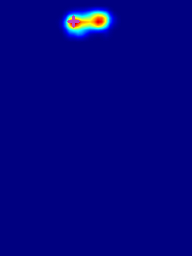}
  \end{subfigure}
  
  \begin{subfigure}{\linewidth}
    \centering
    \includegraphics[width=0.495\linewidth]{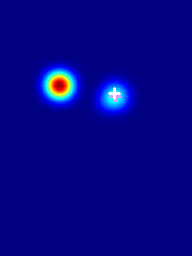}
    \hfill
    \includegraphics[width=0.495\linewidth]{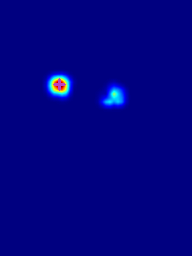}
  \end{subfigure}
  
  \begin{subfigure}{\linewidth}
    \centering
    \includegraphics[width=0.495\linewidth]{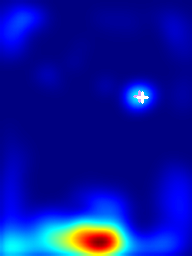}
    \hfill
    \includegraphics[width=0.495\linewidth]{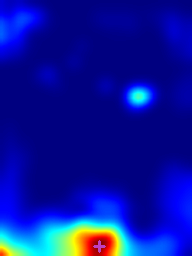}
  \end{subfigure}

    \caption{
    Decoding predicted probability maps throught UDP (left) and expected OKS maximization (right).
    Probability map maximum in white, UDP-refined point in light purple and maximal expected OKS in dark purple.
    These are examples where the decoding does matter as the distribution is multimodal.
    Expected OKS have sharper peaks but predict optimal location globaly in areas with biggest "mass" even though the maximal value could be elsewhere.
    }
    \label{fig:UDP-OKS-bad}
  
\end{figure}

The difference between the UDP and OKS maximization decodings is shown in \cref{fig:UDP-OKS-good,fig:UDP-OKS-bad}. When the predicted probability map is unimodal (which is true for most predicted heatmaps), the difference is negligible. However, if the probability map is multimodal or lacks a clear peak, expected OKS favors areas with greater mass.

\section{Points on the bounding box border}
\label{sec:border-points}

When we evaluated ProbPose qualitatively on the standard COCO dataset, we observed that the ground truth annotations were not always where we expected, particularly in cases where OKS scores worsened the most. Specifically, we noticed that ground truth keypoints near the bounding box border were annotated inside the box but should have been placed outside. It appears that human annotators for the COCO dataset prioritized annotating as many keypoints as possible, even at the cost of accuracy.

\begin{figure}[tb]
  \centering
  \begin{subfigure}{\linewidth}
    \centering
    \includegraphics[width=0.495\linewidth,page=9]{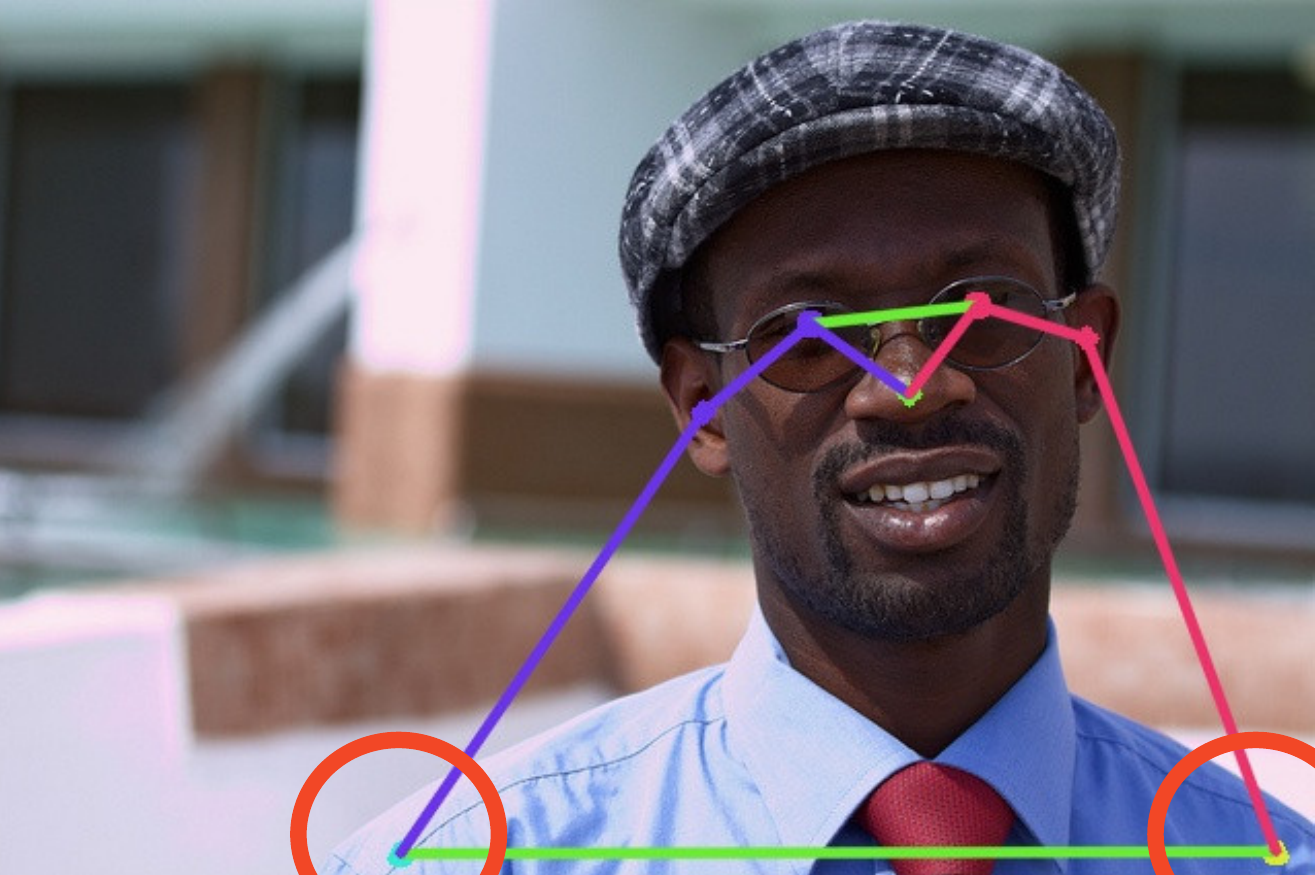}
    \hfill
    \includegraphics[width=0.47\linewidth,page=10]{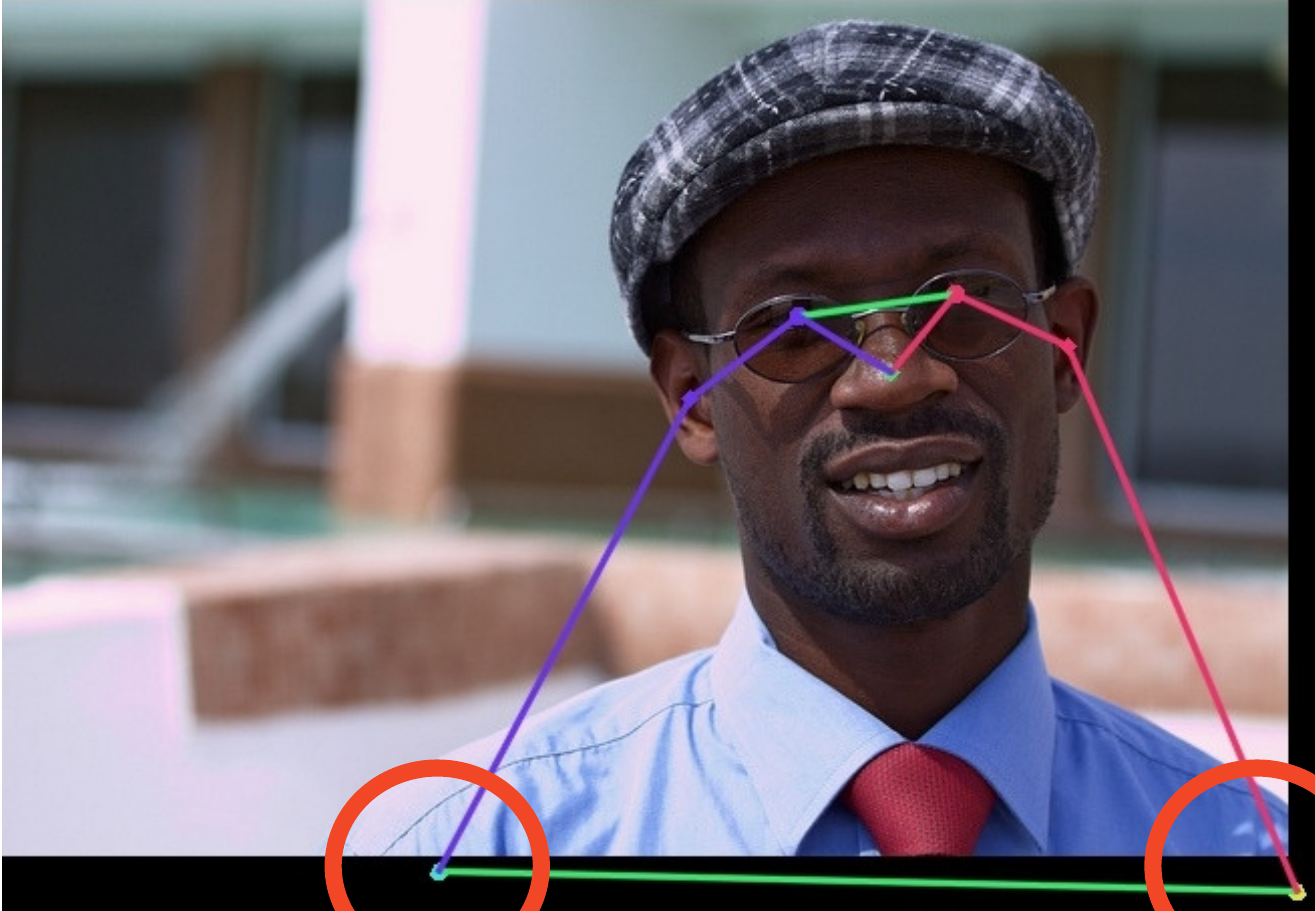}
  \end{subfigure}

  \begin{subfigure}{\linewidth}
    \centering
    \includegraphics[width=0.449\linewidth,page=7]{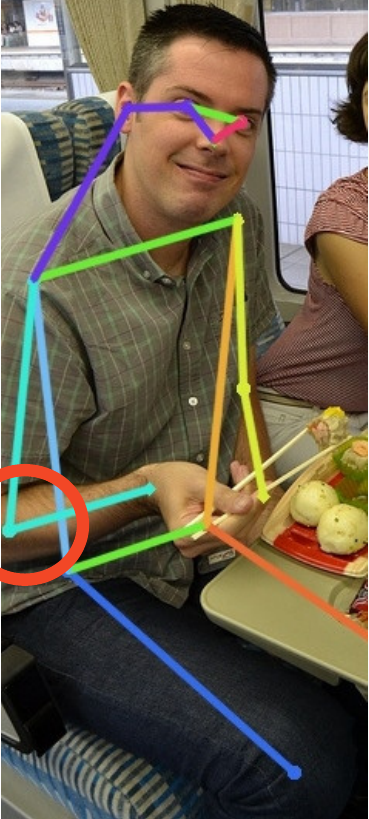}
    \hfill
    \includegraphics[width=0.50\linewidth,page=8]{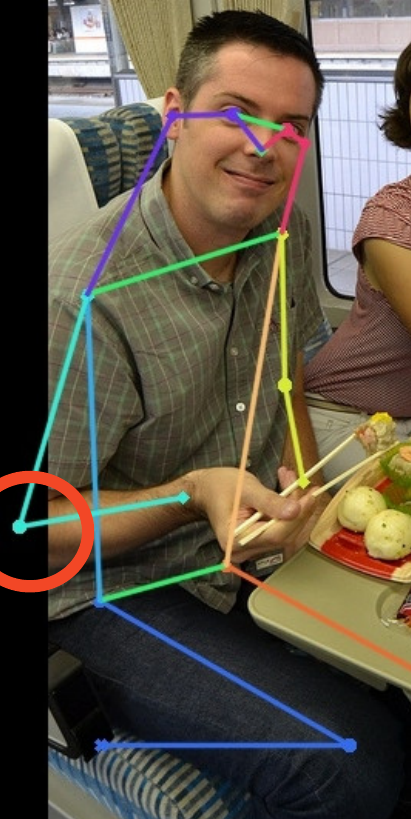}
  \end{subfigure}
  
  \begin{subfigure}{\linewidth}
    \centering
    \includegraphics[width=0.495\linewidth,page=5]{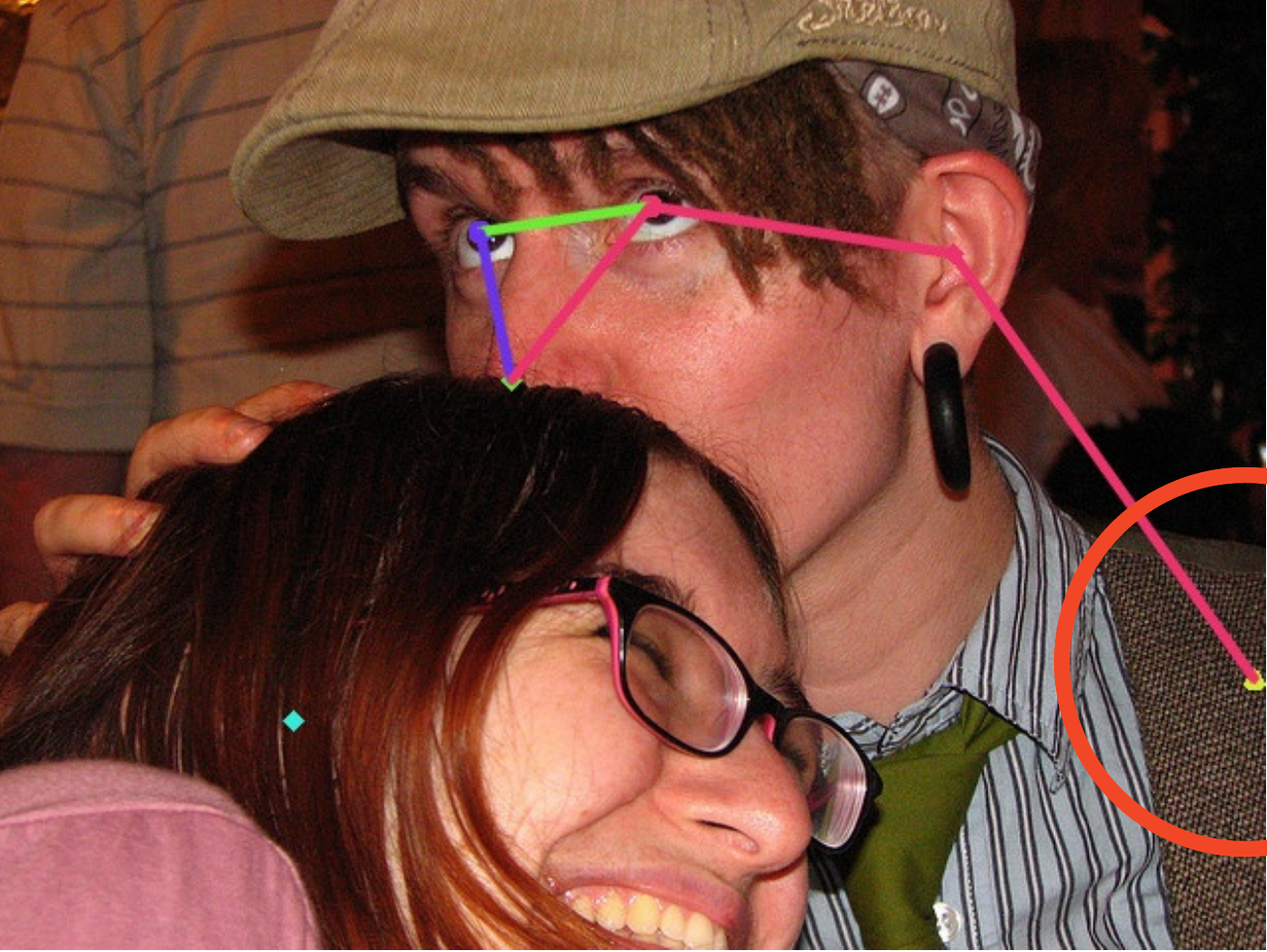}
    \hfill
    \includegraphics[width=0.495\linewidth,page=6]{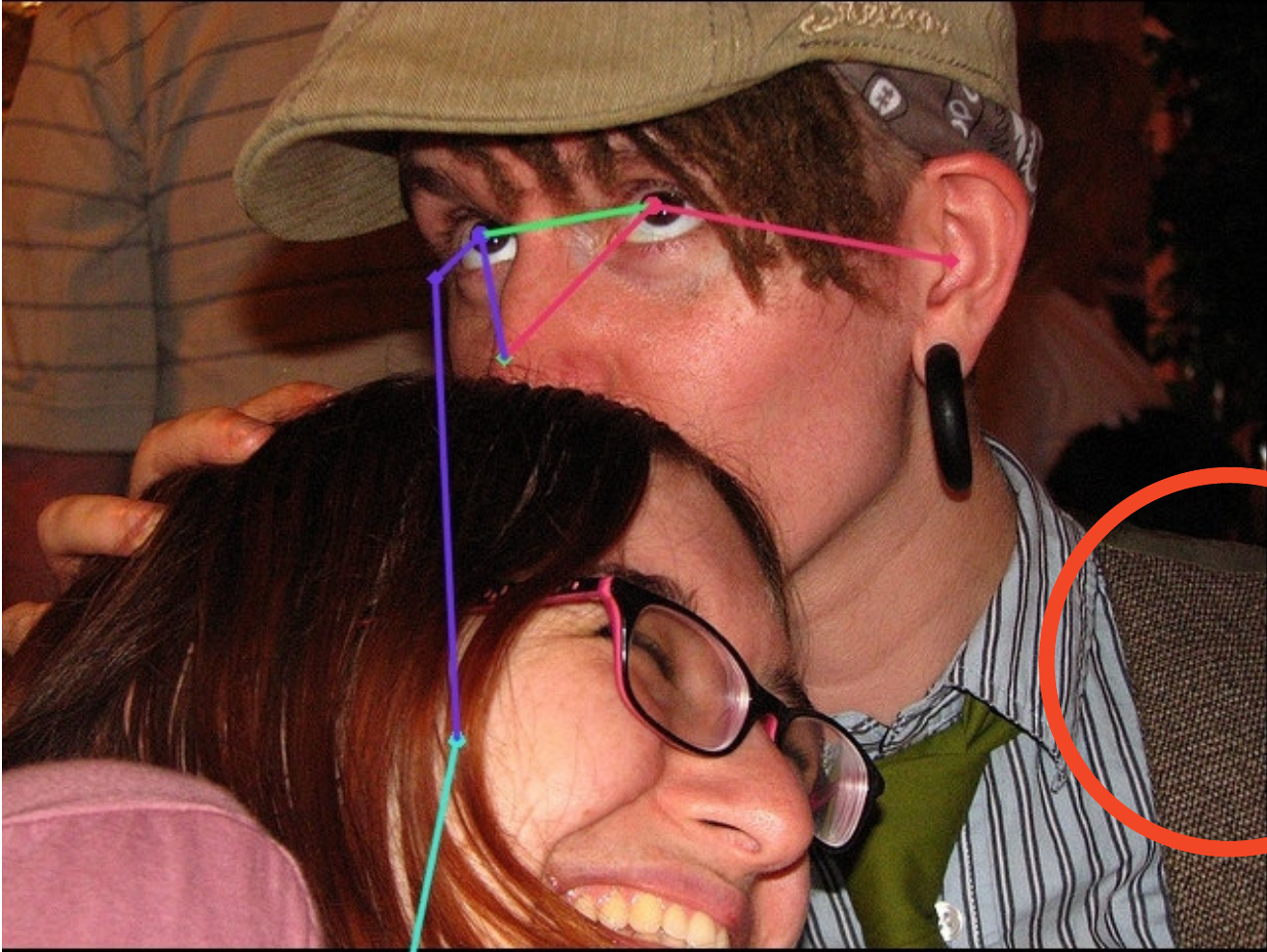}
  \end{subfigure}

  \begin{subfigure}{\linewidth}
    \centering
    \includegraphics[width=0.495\linewidth,page=11]{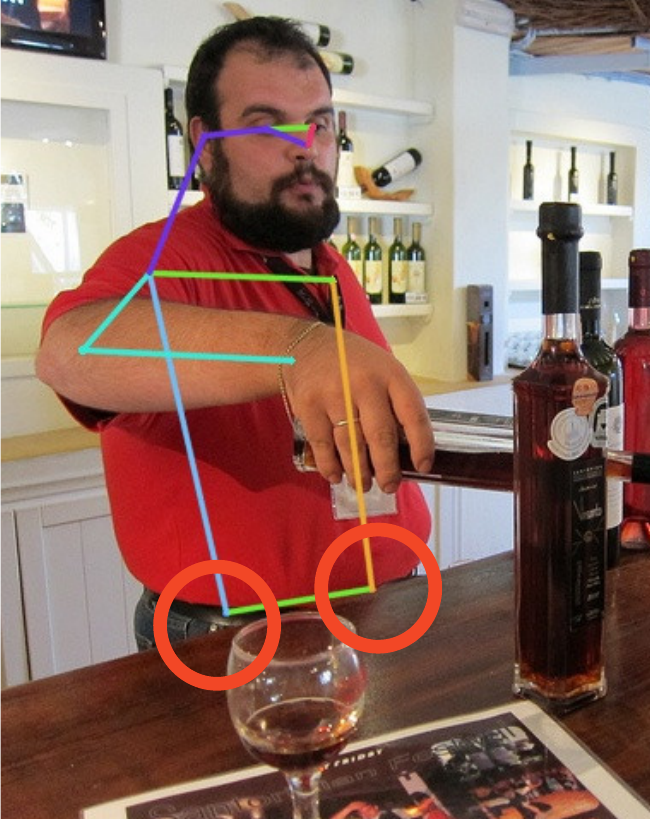}
    \hfill
    \includegraphics[width=0.495\linewidth,page=12]{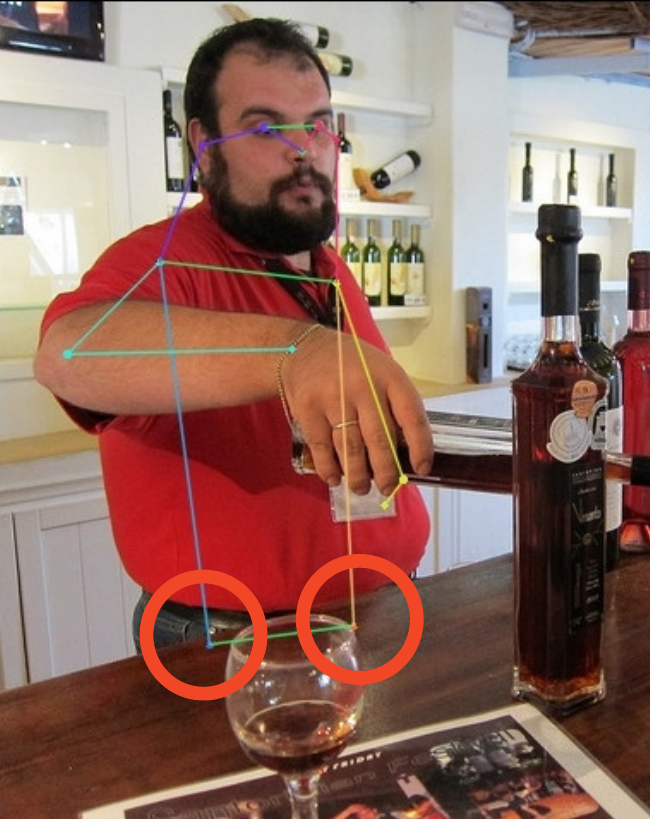}
  \end{subfigure}

    \caption{
    Ground truth annotation (left) vs. ProbPose-s (right) on COCO.
    Images showcasing dubious annotations along the bounding box border where ProbPose gets penalized even though its input seems better than ground truth.
    In the third row, our estimate is missing as we correctly predict it outside of the activation window.
    The problem is not only along the image border as shown in the last row.
    }
    \label{fig:COCO-errors}
  
\end{figure}

Examples of such misannotations are shown in \cref{fig:COCO-errors}. As illustrated in the last row, this issue is not limited to the image border but also occurs along the bounding box border. This supports our hypothesis that the image border behaves as another form of occlusion, as discussed in \cref{sec:keypoints-analysis}.

ProbPose demonstrates that training with crop data augmentation can help mitigate the impact of these incorrect annotations in COCO. However, we did not find an easy and automated solution to fix this issue in the evaluation set. Ignoring points near the bounding box border during the evaluation showed that ProbPose performs even better, but this approach also excludes many correctly annotated and presumably challenging keypoints. For future research involving keypoints near and beyond the bounding box border, manual reannotation may be necessary to address these errors. 

\end{document}